\pdfoutput=1
\PassOptionsToPackage{dvipsnames}{xcolor}
\documentclass[11pt]{article}

\usepackage[final]{acl}

\usepackage{times}
\usepackage{latexsym}

\usepackage[T1]{fontenc}
\usepackage[utf8]{inputenc}

\usepackage{microtype}

\usepackage{inconsolata}

\usepackage[dvipsnames]{xcolor} 
\usepackage{graphicx}

\usepackage{booktabs}
\usepackage{multirow}
\usepackage{rotating}
\usepackage{subcaption}
\usepackage{amsmath,amssymb}
\usepackage{enumitem}
\usepackage{pifont}% http://ctan.org/pkg/pifont
\newcommand{\xmark}{\ding{55}}%

\title{It's All About In-Context Learning! \\Teaching Extremely Low-Resource Languages to LLMs}
\author{Yue Li, Zhixue Zhao \and Carolina Scarton \\
        Department of Computer Science, University of Sheffield, UK\\
        \texttt{\{yli381,zhixue.zhao,c.scarton\}@sheffield.ac.uk}}

\begin{document}
\maketitle
\begin{abstract}

Extremely low-resource languages, especially those written in rare scripts, as shown in Figure~\ref{fig:language}, remain largely unsupported by large language models (LLMs). This is due in part to compounding factors such as the lack of training data. This paper delivers the first comprehensive analysis of whether LLMs can acquire such languages purely via in-context learning (ICL), with or without auxiliary alignment signals, and how these methods compare to parameter-efficient fine-tuning (PEFT). We systematically evaluate 20 under-represented languages across three state-of-the-art multilingual LLMs. 
%Our results reveal that for languages with minimal exposure during pre-training and poor script coverage, small-scale fine-tuning yields negligible gains due to overfitting and tokenization inefficiencies. In contrast, zero-shot ICL enhanced with simple sentence- or word-level alignment can unlock unexpectedly strong performance, often surpassing fine-tuning and even comparable to continued pre-training approaches. We further demonstrate that few-shot ICL and PEFT are only beneficial for languages where the model already exhibits a modicum of capability. 
% Low-resource languages often suffer from a lack of data availability for pre-training or continue pre-training large language models (LLMs) for language adaptation.
% In the context of using LLMs for downstream tasks, this paper explores guiding an LLM to \textit{learn} low-resource languages through in-context learning (ICL) enhanced with auxiliary signals, such as sentence-level or word-level language alignment. The proposed approach is analysed and compared to parameter-efficient fine-tuning (PEFT) across 20 extremely low-resource languages, including five written with rare scripts (i.e., \textit{Dzongkha}, \textit{Santali}, \textit{Nko}, \textit{Tamasheq} and \textit{Tigrinya}). %and 15 in relatively common scripts such as Latin, Arabic, and Cyrillic. 
Our findings highlight the limitation of PEFT when both language and its script are extremely under-represented by the LLM. In contrast, zero-shot ICL with language alignment is impressively effective on extremely low-resource languages, while few-shot ICL or PEFT is more beneficial for languages relatively better represented by LLMs.
%where LLMs already exhibit some capability. 
For LLM practitioners working on extremely low-resource languages, we summarise guidelines grounded by our results on adapting LLMs to low-resource languages, e.g., avoiding fine-tuning a multilingual model on languages of unseen scripts.

%increasing LLMs' language support on downstream tasks. %low-resource language speakers.
%%%%%%%%
%This paper explores %the possibility of 
%replacing fine-tuning with in-context learning (ICL) guiding a model to learn a brand new language. Specifically, we investigate %if it is possible to 
%whether LLMs can be efficiently adapt to rare (and very likely unseen) languages for downstream tasks, with a focus on the trade-off between resource and performance. %We extensively study 
%ICL and parameter-efficient fine-tuning strategies are extensively studied for five extremely low-resource languages with rare scripts (i.e., \textit{Dzongkha}, \textit{Santali}, \textit{NKo}, \textit{Tamasheq} and \textit{Tigrinya}), as well as 15 low-resource languages written in relatively common scripts such as Latin, Arabic, and Cyrillic. Our study reveals the notable difficulty of adaption with fine-tuning on certain rare languages caused by xxxx, while zero-shot ICL with xxx could bring significant improvements. Something about low-resource languages with relatively common scripts. Overall, our findings offer practical insights and guidance for practitioners when adapting LLMs to rare and underrepresented languages under resource constraints.
%%%%%%
\end{abstract}

\section{Introduction}
%[trim={left bottom right top},clip]
\begin{figure}[ht!]
\centering
\includegraphics[width=.4\textwidth]{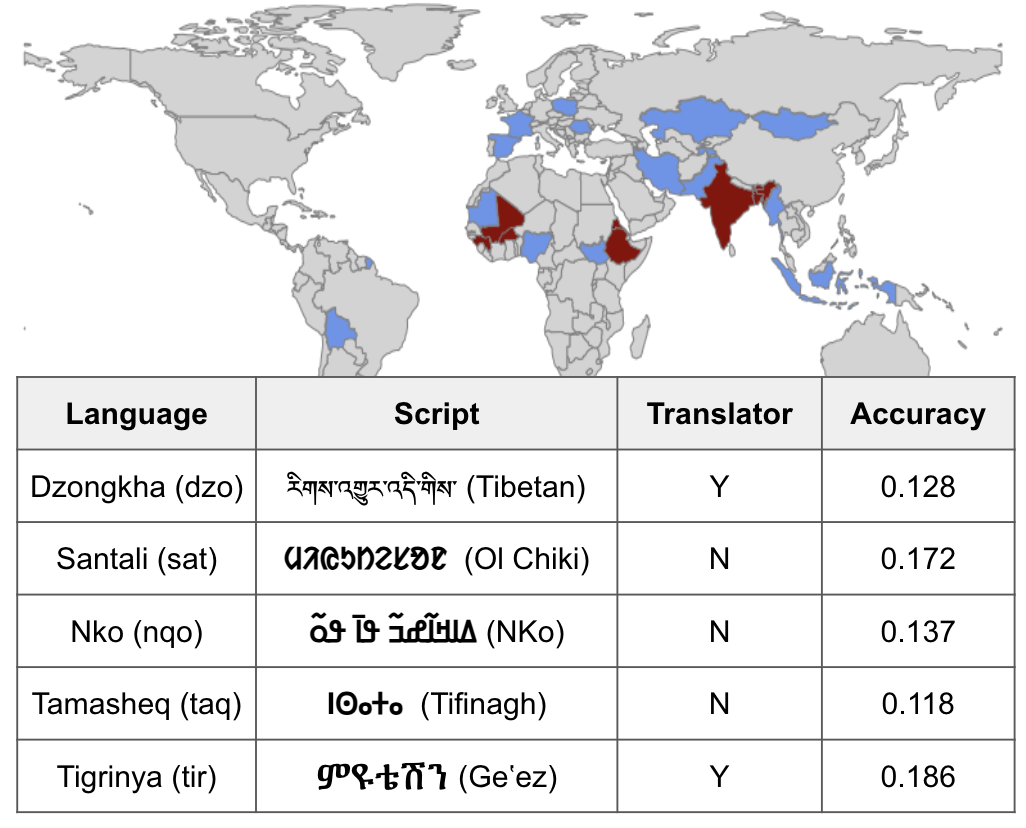}
 \caption[Caption for LOF]{Regional distribution of the languages studied in this paper. \textcolor{red}{Red} 
 %dots \textcolor{red}{\textbullet} 
 denotes the five languages with rare scripts, and \textcolor{blue}{blue} 
 %dots \textcolor{blue}{\textbullet} 
 represents the other 15 languages. \textit{Y} (yes) and \textit{N} (no) denote whether it's supported by Google Translate\protect\footnotemark. \textit{Accuracy} represents performance on topic classification (SIB-200) with DeepSeek (7b) in zero-shot ICL (majority voting = 0.25, English = 0.83).}
 \label{fig:language}
\end{figure}
\footnotetext{\url{https://cloud.google.com/translate/docs/languages}}

Current large language models (LLMs) are typically pre-trained with data in more than 50 languages, offering robust support for high-resource languages, such as German and French~\cite{le2023bloom,grattafiori2024llama,team2023gemini}. However, their coverage of low-resource languages remains limited. Since these languages are often spoken in developing regions, insufficient LLM support risks reinforcing socio-economic disparities and further isolating affected communities~\cite{shen-etal-2024-language,jadhav2024limitations}. 

Extending LLMs to support extremely low-resource languages via continued pre-training \citep{yong-etal-2023-bloom} is possible but often impractical, due to the need for large-scale monolingual corpora and substantial computational resources \citep{joshi-etal-2020-state}. Although LLM support can be achieved for downstream tasks via resource-efficient training, such as parameter-efficient fine-tuning (PEFT), % and zero- or few-shot in-context learning (ICL).
it still requires a non-trivial amount of labeled data. Therefore, with recent advances in in-context learning (ICL), we ask \textbf{whether LLMs can \textit{learn} new languages purely through ICL} \citep{yong-etal-2023-prompting,zhang-etal-2024-teaching,cahyawijaya-etal-2024-llms}. 
Specifically: 
(1) Is ICL alone sufficient to enable LLMs learn extremely low-resource or entirely unseen languages? 
(2) Can auxiliary signals in the prompt be useful enabling or improving ICL? 
(3) ICL or PEFT, which one is a better choice for learning a new language?
%prior work \citep{razumovskaia2024analyzing}?

\begin{table*}[t]
\centering

\scalebox{0.75}{
\begin{tabular}{ll|ccc|cc}
\hline
\multirow{2}{4em}{\textbf{Method}} & & \multicolumn{3}{c|}{\textbf{Data}} & \multirow{2}{3em}{\textbf{Training}} & \multirow{2}{4em}{\textbf{Annotator}} \\
& & \textbf{EN} & \textbf{Target} & \textbf{Other} \\
\hline
\multirow{4}{6em}{Zero-shot ICL} & baseline & \textcolor{OrangeRed}\xmark & \textcolor{OrangeRed}\xmark & \textcolor{OrangeRed}\xmark & \textcolor{OrangeRed}\xmark & \textcolor{OrangeRed}\xmark\\
& sentence-level alignment  & \textcolor{ForestGreen}{\checkmark}($k$) & \textcolor{ForestGreen}{\checkmark}($k$) & \textcolor{OrangeRed}\xmark & \textcolor{OrangeRed}\xmark & \textcolor{ForestGreen}{\textit{translate}}\\
& word-level alignment & \textcolor{OrangeRed}\xmark & \textcolor{OrangeRed}\xmark & \textcolor{ForestGreen}{\textit{dict.}} & \textcolor{OrangeRed}\xmark & \textcolor{OrangeRed}\xmark\\
& word-level translation & \textcolor{OrangeRed}\xmark & \textcolor{OrangeRed}\xmark & \textcolor{ForestGreen}{\textit{dict.}} & \textcolor{OrangeRed}\xmark & \textcolor{OrangeRed}\xmark \\
\hline
\multirow{2}{6em}{Few-shot ICL} & demonstration in target language & \textcolor{OrangeRed}\xmark & \textcolor{ForestGreen}\checkmark($k$) & \textcolor{OrangeRed}\xmark & \textcolor{OrangeRed}\xmark &\textcolor{ForestGreen}{\textit{label}}\\
& demonstration with alignment & \textcolor{ForestGreen}\checkmark($k$) & \textcolor{ForestGreen}\checkmark($k$) & \textcolor{OrangeRed}\xmark & \textcolor{OrangeRed}\xmark & \textcolor{ForestGreen}{\textit{label+translate}}\\
\hline
\multicolumn{2}{l|}{Parameter efficient fine-tuning} & \textcolor{OrangeRed}\xmark & \textcolor{ForestGreen}\checkmark($N$) & \textcolor{OrangeRed}\xmark & \textcolor{ForestGreen}{\checkmark} & \textcolor{ForestGreen}{\textit{label}}\\
\hline
\end{tabular}
}
\caption{List of methods used in this paper and resources they may rely on: (1) Data: in-domain data in English (EN) or target language (Target), or a dictionary (dict.) for word translation; $k$ denotes $k$-shot examples, and
$N$ denotes the full training data ($k\ll N$); (2) Training: whether model parameters are updated; (3) Annotator: whether native speakers are needed to \textit{label} the data, or \textit{translate} data to English to enable alignment.}
\label{tab:methods}
\end{table*}

In this study, we consider 20 low-resource languages, including five extremely low-resource ones (Figure \ref{fig:language}) and 15 written in Latin, Arabic, or Cyrillic scripts (referred to as \textit{target} languages\footnote{represented by language code ISO 639-3 (in Table \ref{tab:language full list})}). %Besides ICL alone, 
ICL with auxiliary signals (i.e., class category, language alignment) is explored. Our setup (Table~\ref{tab:methods}) includes few-shot ICL, sentence-level alignment of unlabelled or labelled examples in zero-shot or few-shot ICL, and word-level alignment for the target language in zero-shot ICL. %We analyse their effectiveness comparing with the baseline zero-shot ICL where LLMs are only prompted with the English task description and target-language input. We also compare the effective ICL approaches with fine-tuning, and provide useful suggestions for practitioners working on different kinds of low-resource languages.

To our knowledge, this is the first study to systematically analyse whether ICL can enable LLMs to \textit{learn} extremely low-resource languages. Our main findings are:

\begin{itemize}[leftmargin=*]
    
    \item In contrast to prior work \citep{razumovskaia2024analyzing}, small-scaled fine-tuning is generally ineffective when a language and its script are highly under-represented or entirely absent from both the tokenizer and pre-training data (e.g., \texttt{sat}, \texttt{nqo} and \texttt{taq} on DeepSeek).
    
    \item In such cases, zero-shot ICL with language alignment yields substantial gains, potentially surpassing vocabulary extension on multilingual pre-trained language models (PLMs) through continue pre-training.

    \item Zero-shot ICL with language alignment is especially effective  for languages with minimal LLM support, often exceeding few-shot ICL and performing comparably to, or better than, fine-tuning. %which are limitedly supported by LLMs, often better than few-shot ICL and even comparable to or better than fine-tuning. 

    \item Few-shot ICL and especially PEFT perform best for low-resource languages for which LLMs exhibit a certain level of support.

\end{itemize}

\section{Related Work}

\paragraph{Language Adaptation in Pretraining}

Continued pretraining LLMs on a monolingual corpus in a target language is a common strategy to extend support to languages not (well) represented in the original pretraining data, also enhancing ICL performance in the target language~\citep{yong-etal-2023-bloom}. Various methods have been explored for training efficiency \citep{zhang2021cpm,yong-etal-2023-bloom,cui2023efficient}, vocabulary and tokeniser adaptation \citep{yamaguchi-etal-2024-empirical,balachandran2023tamil,cui2023efficient,larcher2023cabrita} and data efficiency \citep{yamaguchi2024can,shaham-etal-2024-multilingual,kurz2024investigating}. However, the effectiveness of these pretraining-based methods often depends on the availability of large-scale training data, an assumption that does not hold for extremely low-resource languages in real-world scenarios.

\paragraph{Adapting LLMs to Low-Resource Languages for Downstream Tasks} ICL with different strategies has been explored to improve LLMs' adaptation to low-resource languages, 
including techniques such as code-switching \citep{yong-etal-2023-prompting,schlicht2025llms}, demonstration example selection \citep{winata-etal-2022-cross,zhang-etal-2024-teaching,tanwar-etal-2023-multilingual}, prompt format optimization \citep{zhang-etal-2023-dont,cahyawijaya-etal-2024-llms}, machine translation \citep{bandarkar-etal-2024-belebele}, and dictionary-based prompting \citep{lu-etal-2024-chain,zhang-etal-2024-teaching}. Another promising direction is PEFT, which has demonstrated superior performance with computational costs comparable to few-shot ICL \citep{liu2022fewshot}. 
However, most existing studies focus on languages that are: (1) relatively high-resource (e.g., German); (2) low-resource but written in widely supported scripts (e.g., Zhuang in Latin script \citep{zhang-etal-2024-teaching}); and (3) written in rare scripts but already included in model pre-training \citep{razumovskaia2024analyzing}. 
Consequently, how to effectively adapt LLMs to extremely low-resource languages such as \texttt{nqo} (Figure~\ref{fig:language}) is still unclear. 

%Therefore, %However, adapting LLMs to languages that are severely under-represented in both tokenizer and model potentially raises new challenges, such as long input sequences and no tools available for machine translation. 

\section{Learning Extremely Low-Resource Languages}\label{sec:learning}
%In this section, we compare two main ICL  (Sec.~\ref{}).....

Table \ref{tab:methods} summarises the experimented approaches and assumed available data resources. Standard cross-lingual transfer that aims to improve and then transfer task knowledge from English is not considered in this study (i.e, fine-tuning with English data or few-shot ICL with English examples), as the LLMs have already demonstrated high accuracy on English in zero-shot ICL (Table \ref{tab:PEFT full results})\footnote{Machine translating target-languages into English is not considered, since our aim is to teach low-resource languages to LLMs. Reliable machine translators are also not available for 3 out of 5 rare-script languages. In practice, developing a high-quality machine translator is significantly more expensive than creating the data resources we consider here.}. 

\paragraph{Baseline} The vanilla zero-shot ICL when the LLMs are prompted only with task description and the target language input $tgt$. We use English task description as %it could improve ICL performance on non-English data and 
it has been widely adopted showing improvements in ICL performance \citep{zhang-etal-2023-dont,razumovskaia2024analyzing}. The prompt format is: [<task description> + <input$^{tgt}$>]. 

\paragraph{Zero-Shot ICL with alignment} we experiment with adding word- or sentence-level alignment between English and a target language in the prompt, without providing labelled examples. 
\begin{itemize}[leftmargin=*]
    \item \textit{Word-level alignment}: We provide a translation for each word in the target-language input using a dictionary, inspired by prior work on machine translation \citep{zhang-etal-2024-teaching,lu-etal-2024-chain}. The prompt format for an input$^{tgt}$ with $N$ words \{w$^{tgt}_1$, w$^{tgt}_2$, ...,  w$^{tgt}_N$\} is: [<task description> + <input$^{tgt}$> + <w$^{tgt}_1$ means w$^{eng}_1$ in English; ...; w$^{tgt}_N$ means w$^{eng}_N$ in English>]. We use the NLLB translator\footnote{\url{https://huggingface.co/facebook/nllb-200-3.3B}} \citep{costa2022no} to create the dictionaries following \citet{lu-etal-2024-chain}. For languages not supported by NLLB (\texttt{nqo}, \texttt{sat}\footnote{NLLB repeats the same word without stopping}, and \texttt{min}), we
    train the word alignment tool \textit{fast\_align} \citep{dyer-etal-2013-simple} to simulate a high-quality dictionary (See Appendix \ref{app:imp}).

    \item \textit{Word-level translation}: We directly concatenate the English word translations in their orders in target languages as the ``English'' translation (i.e., input$^{eng'}$ = concat (w$^{eng}_1$, ..., w$^{eng}_N$)), and prompt LLMs with: [<task description> + input$^{eng'}$].

    \item \textit{Sentence-level alignment}: Assuming there is a limited $k$ number of parallel in-domain unlabelled sentences in English \{s$^{eng}_1$,...,s$^{eng}_k$\} and target language \{s$^{tgt}_1$,...,s$^{tgt}_k$\}. The prompt format is: [<target language: s$^{tgt}_1$; English: s$^{eng}_1$; ...; target language: s$^{tgt}_k$; English: s$^{eng}_k$> + <task description> + <input$^{tgt}$>]. We select the target-language example sentences from the training data through random sampling or BM25 \citep{robertson2009probabilistic}.
    
    %based on the target languages 

\end{itemize}

\paragraph{Few-Shot ICL} %We explore the following two settings, 
Assuming there is a limited number of labelled in-domain data samples in the target language or English, demonstration examples from the training data are retrieved by BM25, inspired by \citet{zhang-etal-2024-teaching}. %illustrate BM25 effectiveness over random sampling in few-shot ICL for low-resource languages).

\begin{itemize}[leftmargin=*]

    \item \textit{Demonstration in target language}: We prompt LLMs with $k$-shot demonstration examples in the target language D$^{tgt}_1$, ..., D$^{tgt}_k$. 
    %(D$^{tgt}_i$ = s$^{tgt}_i$ with label $i$). 
    The prompt format is [<task description> + <D$^{tgt}_1$, ..., D$^{tgt}_k$> + <input$^{tgt}$>].
    
    \item \textit{Demonstration with alignment}: LLMs are prompted with parallel demonstration examples in both English and target languages: [<task description> + <D$^{tgt}_1$ means D$^{eng}_1$, ..., D$^{tgt}_k$ means D$^{eng}_k$> + <input$^{tgt}$>].
    
\end{itemize}

\paragraph{PEFT} We preliminarily experiment with competitive methods such as LoRA \citep{hu2022lora}, DoRA \citep{10.5555/3692070.3693369} and IA3 \citep{liu2022fewshot}. Same as \citet{yong-etal-2023-bloom}, we found that IA3 is the most effective and efficient approach. Therefore, due to computational constraints, we only experiment with IA3 as a representative of the PEFT methods. We also discuss the comparison with fully fine-tuned multilingual PLMs in Section \ref{sec:compareall}.

%\subsection{Adaptation Strategies}

\subsection{Experimental Setups}\label{sec:experimental}

%\subsection{Target Languages}

%\begin{figure}[ht!]
% \centering
% \includegraphics[width=.45\textwidth]
 %{latex/image/language_2.png}
 %\caption{Examples of 5 low-resource languages with rare scripts explored in this study. The cross mark denotes machine translation that is not supported by Google. Macro-F1 scores on SIB-200 with Deepseek in zero-shot ICL is presented  (random guess baseline = 14.5, English baseline = 79.2).}
 %\label{fig:language}
%\end{figure}

\paragraph{Target Languages} We mainly ground our research on the \textit{SIB-200} seven-way topic classification dataset \citep{adelani-etal-2024-sib}, as it offers parallel training and evaluation data with the broadest multilingual coverage in natural language understanding (NLU) tasks. We also analyse the generalisability of our findings on reading comprehension (i.e., BELEBELE \citep{bandarkar-etal-2024-belebele}), which is a more challenging task than topic classification.
Since most prevalent LLMs do not disclose comprehensive lists of languages present in their pretraining data, we select the low-resource languages for which LLMs exhibit significantly limited capability. We measure the LLMs' capability on each language with \textit{Information Parity} (IP) \citep{tsvetkov-kipnis-2024-information} on the SIB training data. 
Given a text in the target language and its English translation, IP is defined as the ratio between the negative log-likelihood of the target-language text and that of its English counterpart under the same model. A very low IP score indicates that the LLM struggles to represent information in the target language, likely due to limited or no exposure during pretraining. Based on this criterion, we select the languages with the average lowest IP scores across the LLMs we study on. This includes five languages written in relatively rare and distinct scripts (Figure \ref{fig:language}) plus 15 languages using more commonly supported scripts (i.e., seven in Latin, four in Arabic, and four in Cyrillic). For the latter, we select languages that do not have the same linguistic roots as English (Latin script), Modern Standard Arabic (Arabic script) and Russian (Cyrillic script), which are commonly represented in LLMs' training data.
%we focus on the ones with linguistic roots different from English, Modern Standard Arabic and Russian, which are commonly more represented in training data for Latin, Arabic and Cyrillic scripts, respectively. 
The full list of the languages is shown in Table \ref{tab:language full list}. 
%\cass{
%In later section, we also analyse the TBR and TP, both of which tokenization metrics also somehow reflect the how familiar the model is for the given language. Here, we use IP to xxxx.}

\paragraph{Models} We experiment with three recent 
%prevalent 
open-source instruction-tuned LLMs with multilingual ability: \textit{DeepSeek}\footnote{\url{https://huggingface.co/deepseek-ai/deepseek-llm-7b-chat}}, \textit{LlaMA-3.2}\footnote{\url{https://huggingface.co/meta-llama/Llama-3.2-3B-Instruct}} and \textit{Gemma-2}\footnote{\url{https://huggingface.co/google/gemma-2-2b-it}}. Due to computational constraints, their medium-sized variants are considered.

\paragraph{Setups} We adopt accuracy as the evaluation metric following \citet{adelani-etal-2024-sib}. We use greedy search in decoding for the purpose of reproducibility. The prompt template for baseline zero-shot ICL is also used in PEFT for a fair comparison. SIB-200 dataset's official train/dev/test set split is used. The examples included in zero-shot and few-shot ICL are retrieved from the training data. 
As pre-processing for BM25, only white-space splitting is applied.
The hyper-parameter tuning and training details for IA3 are included in Appendix \ref{app:imp}.

\section{Results}

%Section \ref{sec:ft} discusses 
%the limitations of PEFT. Sections \ref{sec:zeroicl} and \ref{sec:fewicl} present the effectiveness of ICL with different alignment settings.
%Section \ref{sec:compareall} compares the performance gains for all approaches, including fine-tuning, zero- and few-shot ICL.

\subsection{Limitation of Fine-Tuning}\label{sec:ft}

\begin{figure*}[h!]
\centering
\begin{subfigure}{0.23\textwidth}
\includegraphics[width=\textwidth]{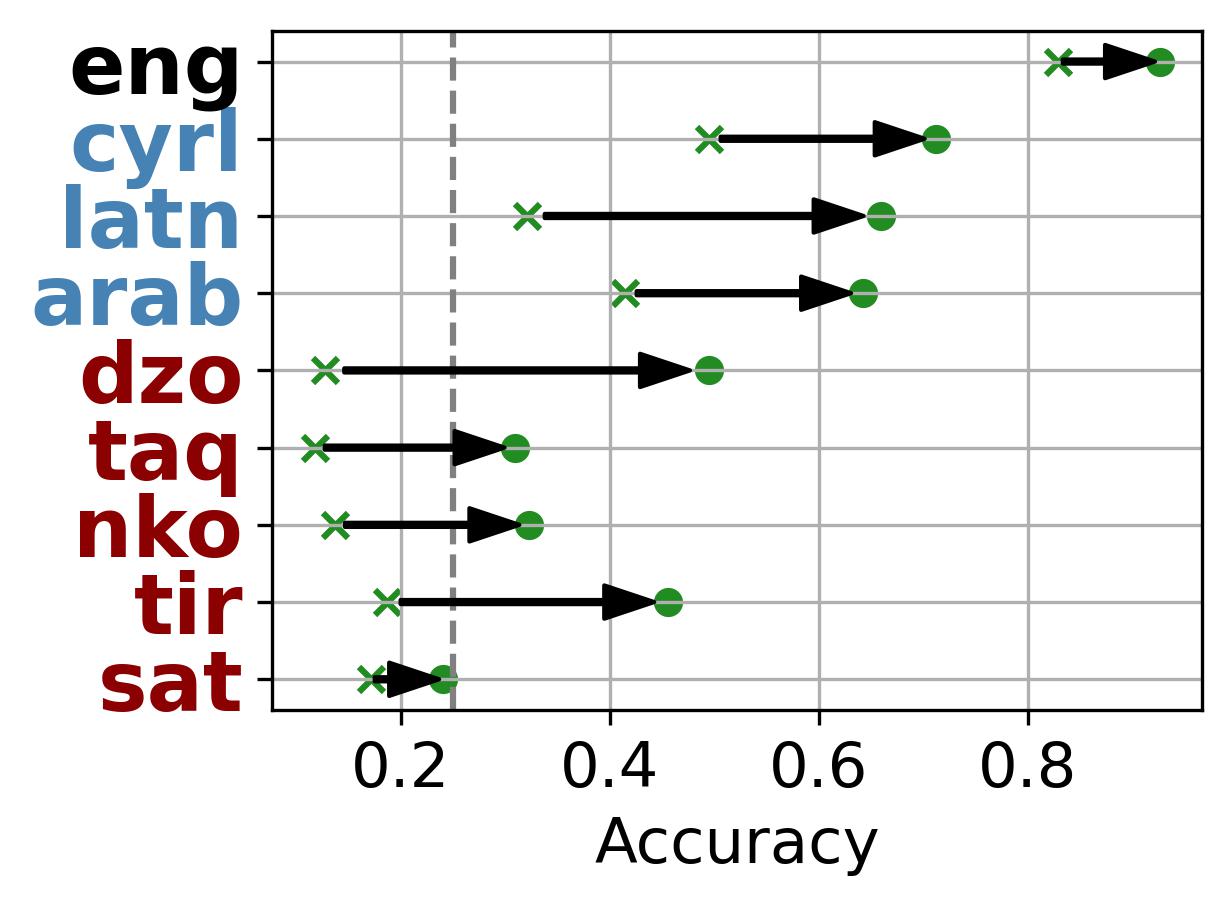}
\caption{DeepSeek}\label{fig:ft-deepseek}
\end{subfigure}
\begin{subfigure}{0.23\textwidth}
\includegraphics[width=\textwidth]{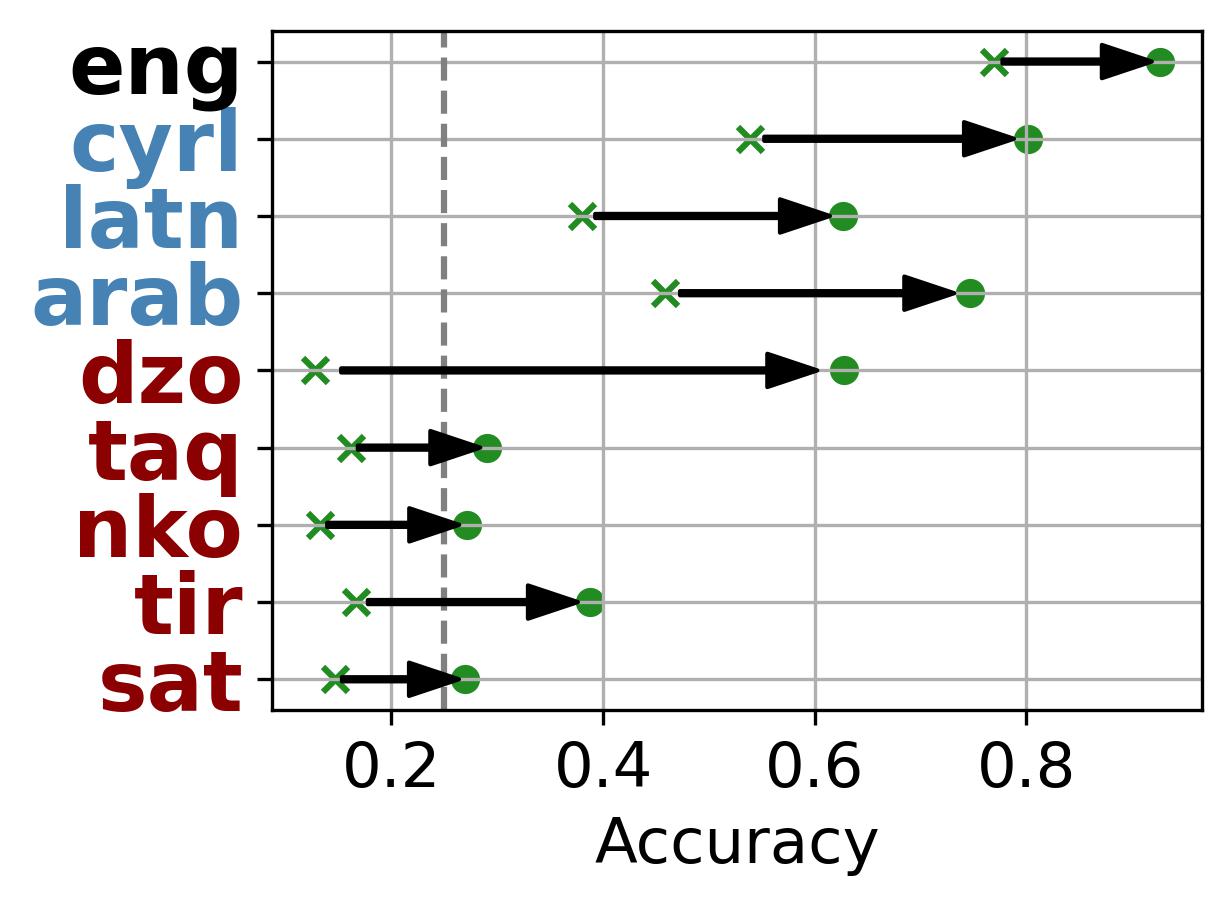}
\caption{LLaMA-3.2}\label{fig:ft-llama32}
\end{subfigure}
\begin{subfigure}{0.23\textwidth}
\includegraphics[width=\textwidth]{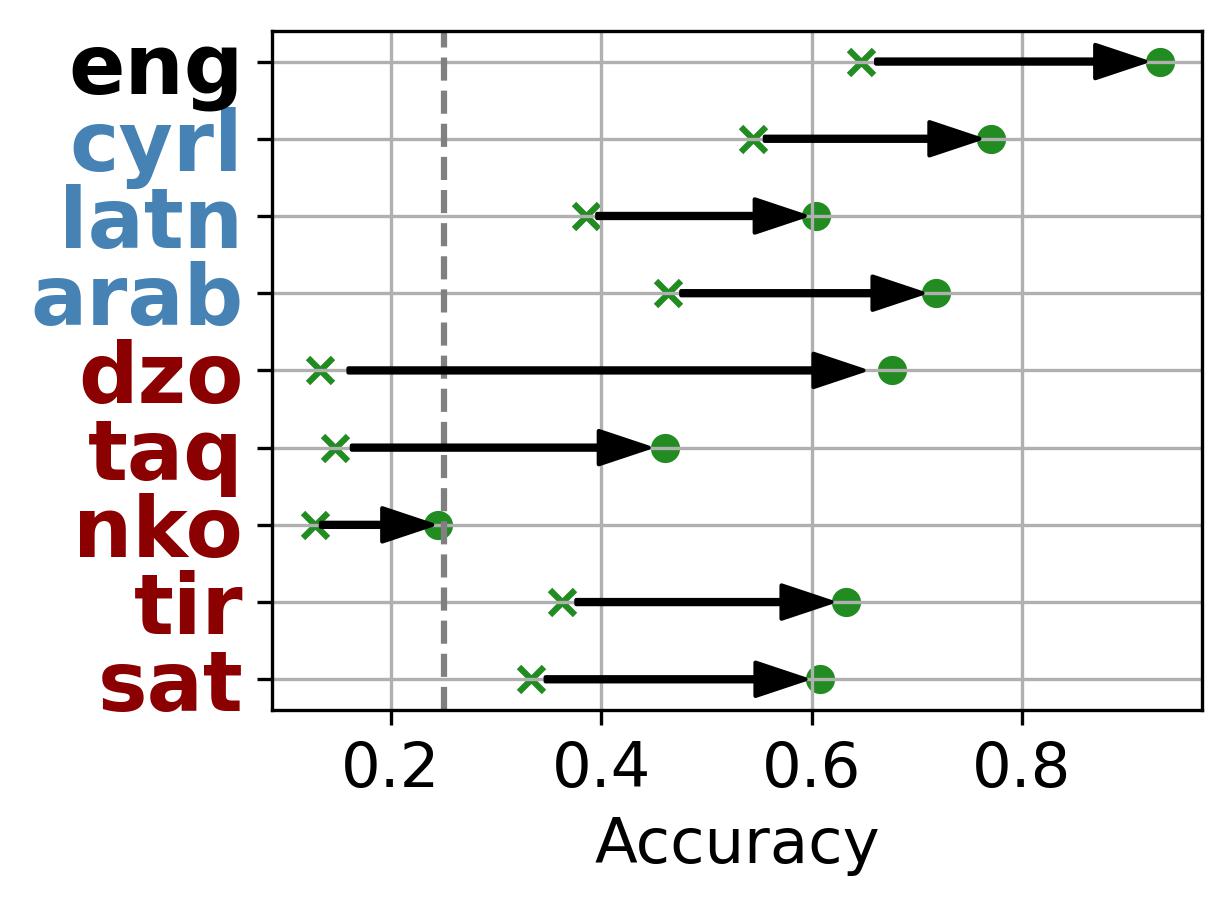}
\caption{Gemma-2}\label{fig:ft-gemma2}
\end{subfigure}
\caption{Accuracy improvement from baseline zero-shot ICL (denoted as \textcolor{ForestGreen}{$\times$}) to PEFT (denoted as \textcolor{ForestGreen}{$\bullet$}). The performance over languages in Latin (\textit{latn}), Arabic (\textit{arab}), and Cyrillic (\textit{cyrl}) scripts is averaged. The performances of English (eng) and the majority voting baseline (\textit{vertical dashed line, accuracy = 0.25}) are for reference.}
\label{fig:ft-results}
\end{figure*}

\paragraph{Fine-tuning Improvement Disparity} Figure \ref{fig:ft-results} illustrates the performance improvement after fine-tuning with the training data in the target language. In most cases, fine-tuning leads to enhanced performance, although the degree of improvement varies notably. For \textcolor{blue}{low-resource languages using common scripts},
%(i.e., Latin, Arabic and Cyrillic), 
accuracy scores can rise to more than 0.6 on average, resulting in an acceptable performance (full results in Table \ref{tab:PEFT full results}). In contrast, results on the five languages written in rare scripts are inconsistent. For instance, while DeepSeek performs worse than majority voting on all of these five languages in the baseline zero-shot ICL setting, PEFT raises the accuracy scores of \texttt{dzo} and \texttt{tir} to above 0.45. In contrast, gains for the remaining three languages are rather modest, particularly for \texttt{sat}, which still stays slightly below majority voting. We observe that this discrepancy is due to overfitting, which appears to occur at a very early stage in the fine-tuning for languages showing limited improvement.

\paragraph{Risk of Overfitting and Impact Factors} 
%The observed severe overfitting can generally be attributed to the large capacity of LLMs relative to the limited amount of available training data. 

\begin{figure*}[h!]
\centering
\begin{subfigure}{0.25\textwidth}
\includegraphics[width=1\linewidth]{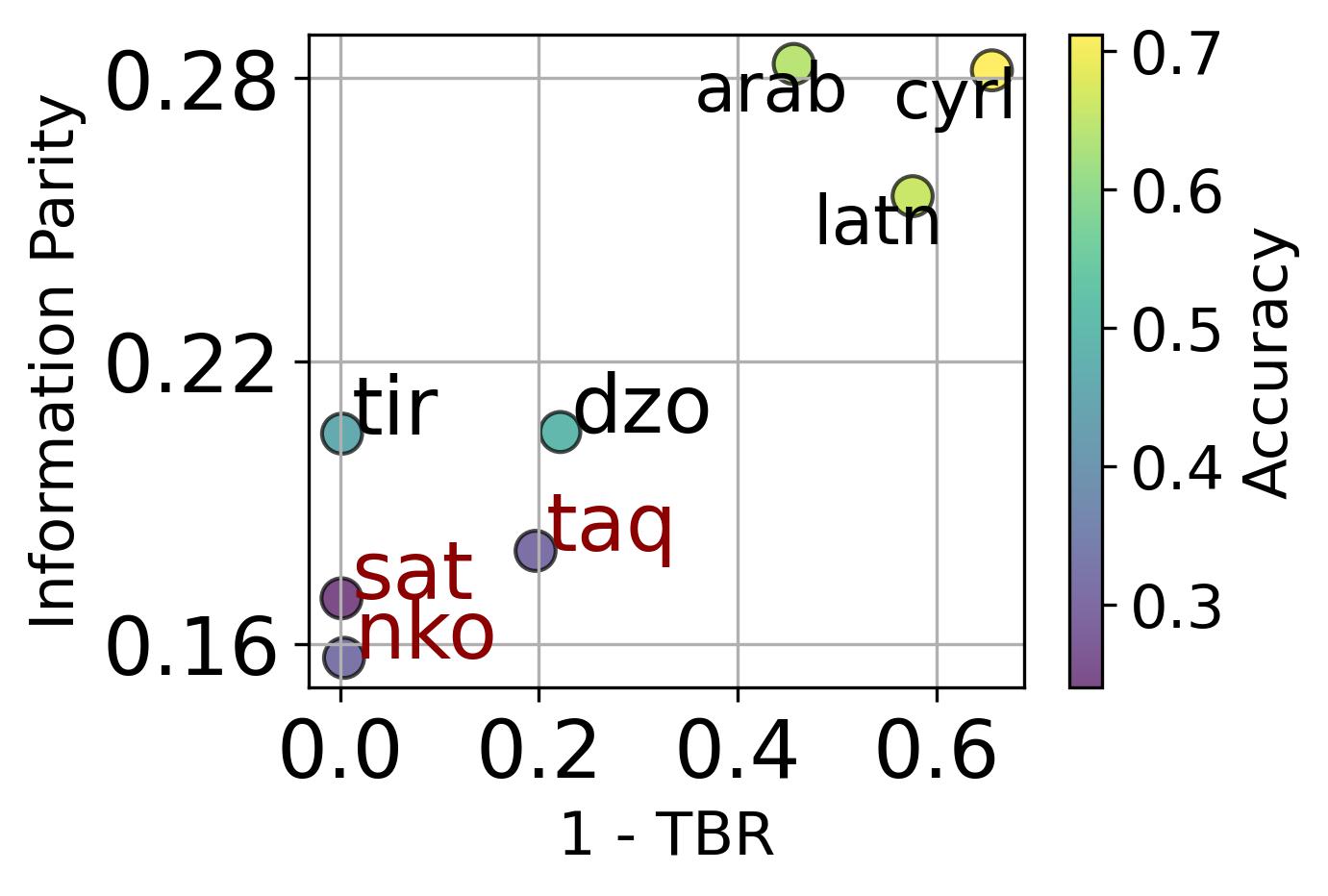}
\caption{DeepSeek}\label{fig:ip-ft-deepseek}
\end{subfigure}
\begin{subfigure}{0.25\textwidth}
\includegraphics[width=1\linewidth]{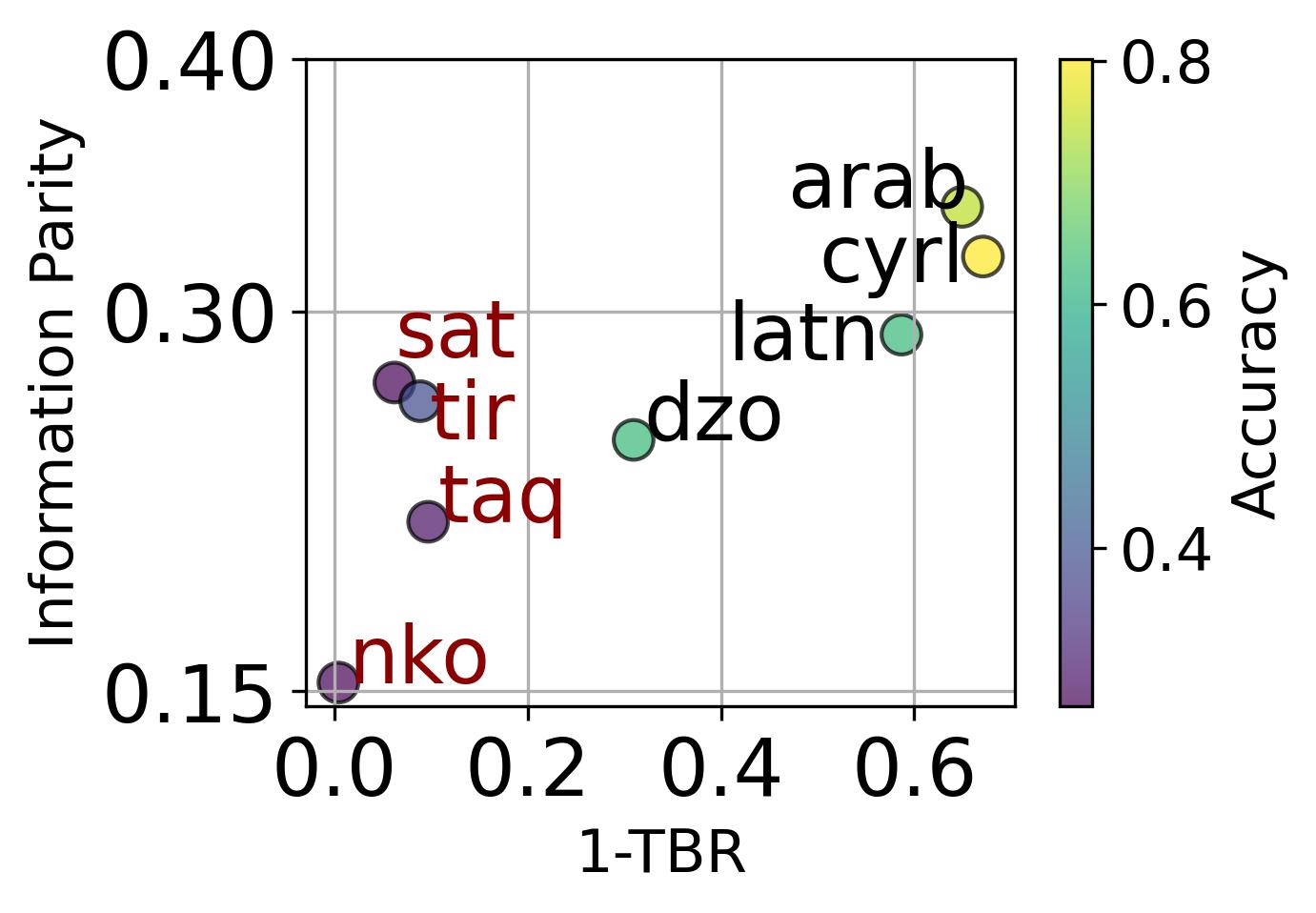}
\caption{LLaMA-3.2}\label{fig:ip-ft-llama32}
\end{subfigure}
\begin{subfigure}{0.25\textwidth}
\includegraphics[width=1\linewidth]{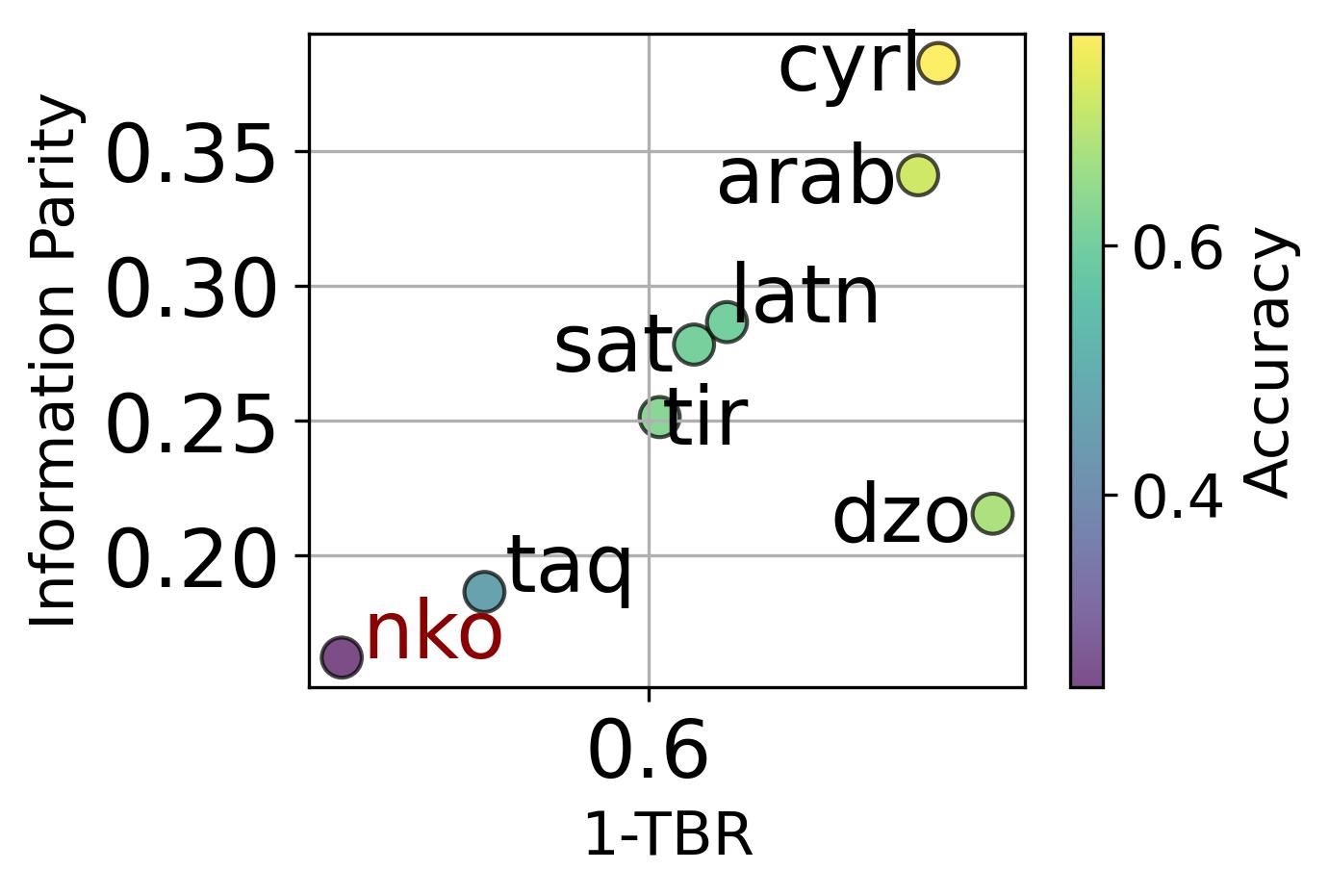}
\caption{Gemma-2}\label{fig:ip-ft-gemma2}
\end{subfigure}
\caption{Correlation between information parity (IP), token-to-byte ratio (TBR) and accuracy score after fine-tuning. For improved visualisation, the x-axis represents ($1-$TBR). The language bubbles close to the left corner denote languages that are under-represented by both tokeniser and model. Bubbles with darker colour denote lower performance after fine-tuning. Names of the languages with PEFT performances lower than 0.4 are marked in red.}
\label{fig:ip-ft-results}
\end{figure*}

To gain more insights on why certain languages suffer more severe overfitting, we analyse: %the following two factors: 

\begin{itemize}[leftmargin=*]
\item \textit{Tokenization efficiency:} Most LLMs' tokenisers, including those used by the three models in our study, adopt byte-level Byte Pair Encoding (BPE) \citep{wang2020neural} or SentencePiece \citep{kudo-richardson-2018-sentencepiece}. When encountering texts in rare scripts not seen during tokeniser training, characters are often segmented into raw bytes, resulting in a vocabulary with drastically reduced effectiveness. For instance, BPE tokenisers based on UTF-8 encoding may end up representing an entire rare-script language using only 256 raw-byte token values \citep{wang2020neural}, limiting the model's ability to learn 
%meaningful or 
generalisable linguistic patterns with small training data~\cite{zhao-aletras-2024-comparing}. 
Tokenisation efficiency for a given text $i$ is measured using \textit{Token-to-Byte Ratio} (TBR) $= \frac{\textit{num}_\text{tokens}^{i}}{\textit{num}_\text{bytes}^{i}}$, where $\textit{num}_\text{bytes}^{i}$ is the number of bytes required to represent the text with the same encoding system used in the tokeniser, and $\textit{num}_\text{tokens}^{i}$ is the number of tokens produced by the LLM's tokeniser. A TBR score close to 1 indicates that the tokeniser is operating nearly at the raw byte level, signalling extremely poor tokenisation. For example, the average TBR for \texttt{sat}'s training data with DeepSeek's tokeniser is 0.99, suggesting that nearly every character is segmented into raw bytes and that DeepSeek significantly lacks meaningful representations for \texttt{sat}. 

\item \textit{Multilingual capability:} %From a transfer learning perspective, 
Fine-tuning is usually more effective when the LLM has already acquired some linguistic competence in the target language during pre-training. IP is used again to estimate the LLMs' capabilities for each target language prior to fine-tuning. As discussed in Section \ref{sec:experimental}, the higher the IP value the more efficiently the LLM represents information provided in the target language. Conversely, a low IP score suggests under-representation of a language.
\end{itemize}

\noindent
Figure \ref{fig:ip-ft-results} shows 
%We compute 
the average TBR and IP scores for each target language in the training data, also presenting their correlations with fine-tuning performance. %, as shown in Figure \ref{fig:ip-ft-results}. 
Languages with high TBR and low IP scores are the ones where fine-tuning tends to encounter more severely overfitting, resulting in very limited generalisation. This suggests that small-scaled fine-tuning 
on downstream tasks is unlikely to be beneficial when a language and its script are highly under-represented or even unseen in both tokeniser training and model pre-training.
This finding also highlights the importance of improving representation of low-resource languages and scripts during pre-training. Even modest improvements in representation, either at tokeniser or model pre-training level, can lead to notably more effective find-tuning adaptation. For instance, although \texttt{tir}, \texttt{sat}, and \texttt{nqo} are nearly entirely tokenised as raw bytes by DeepSeek’s tokeniser (TBR > 0.99), DeepSeek exhibits stronger pre-trained capabilities on \texttt{tir} (higher IP) compared to \texttt{sat} and \texttt{nqo}, translating into more substantial gains from fine-tuning. Similarly, while LLaMA-3.2 shows comparable IP scores for both \texttt{dzo} and \texttt{tir}, \texttt{dzo} benefits from better tokenisation (lower TBR, i.e. higher 1 $-$ TBR), %, and potentially richer vocabulary), 
potentially leading to larger performance improvement after fine-tuning.

\subsection{Alignment in Zero-Shot ICL}\label{sec:zeroicl}

%\begin{figure*}[h!]
%\centering
%\includegraphics[width=.98\textwidth]{latex/image/zeroICL.jpg}
%\caption{Performance on deepseek in zero-shot ICL setting with sentence-level alignment in the prompt. The unlabeled alignment examples (1- or 2-shot) are either retrieved though random sampling or BM-25. Red, green, blue and yellow are used to represent the target languages written in rare scripts, Arabic, Latin and Cyrillic scripts in x-axis, respectively. Languages in each group are ranked based on their baseline zero-shot ICL performances.}
%\label{fig:zeroICL-results}
%\end{figure*}

\subsubsection{Sentence-Level Alignment}
\begin{figure}[h!]
\centering
\begin{subfigure}{0.190\textwidth}
\includegraphics[width=1\linewidth]{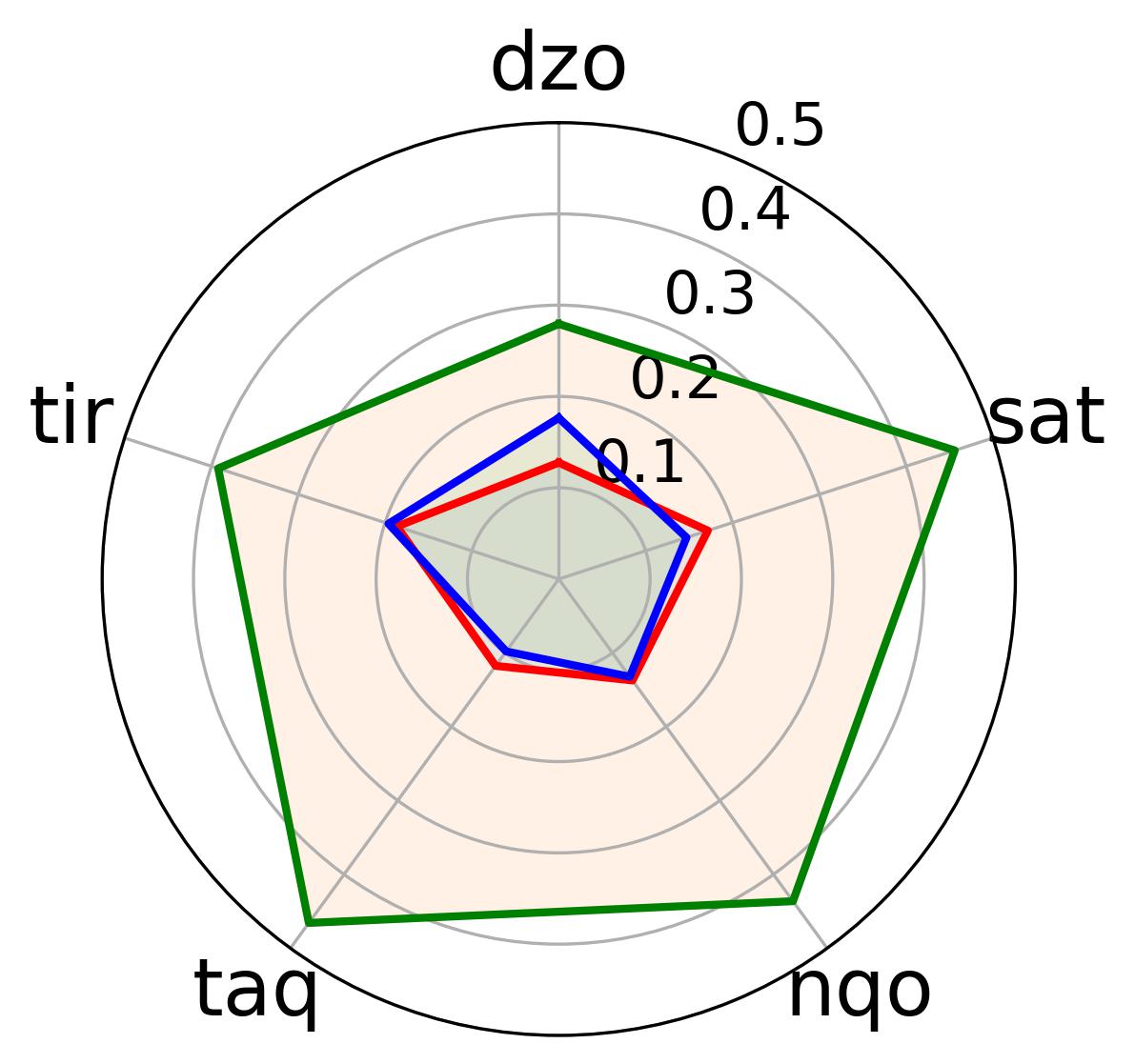}
\caption{Rare Scripts}\label{fig:zero-rare}
\end{subfigure}
\begin{subfigure}{0.190\textwidth}
\includegraphics[width=1\linewidth]{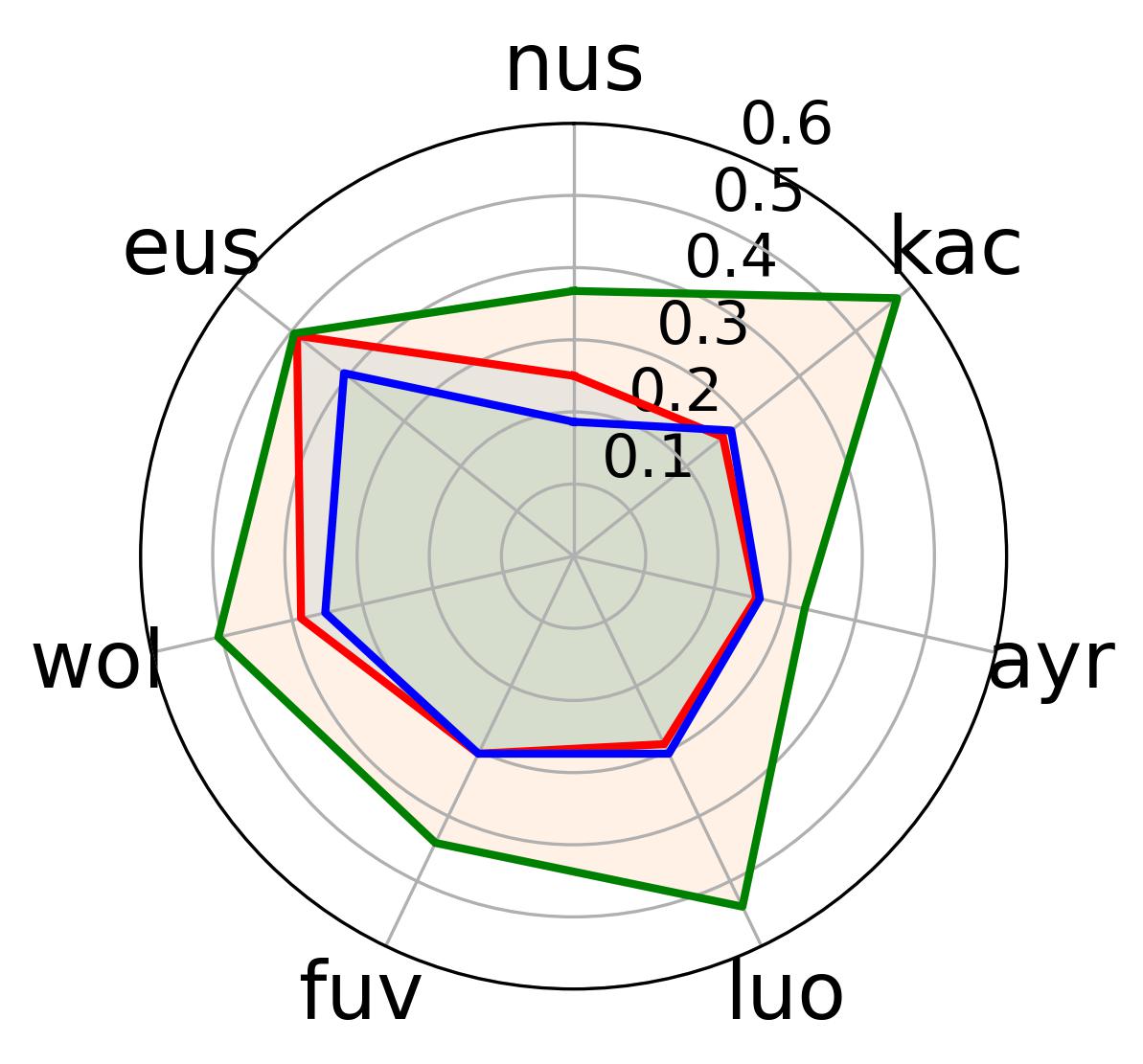}
\caption{Latin}\label{fig:zero-latn}
\end{subfigure}
\begin{subfigure}{0.190\textwidth}
\includegraphics[width=1\linewidth]{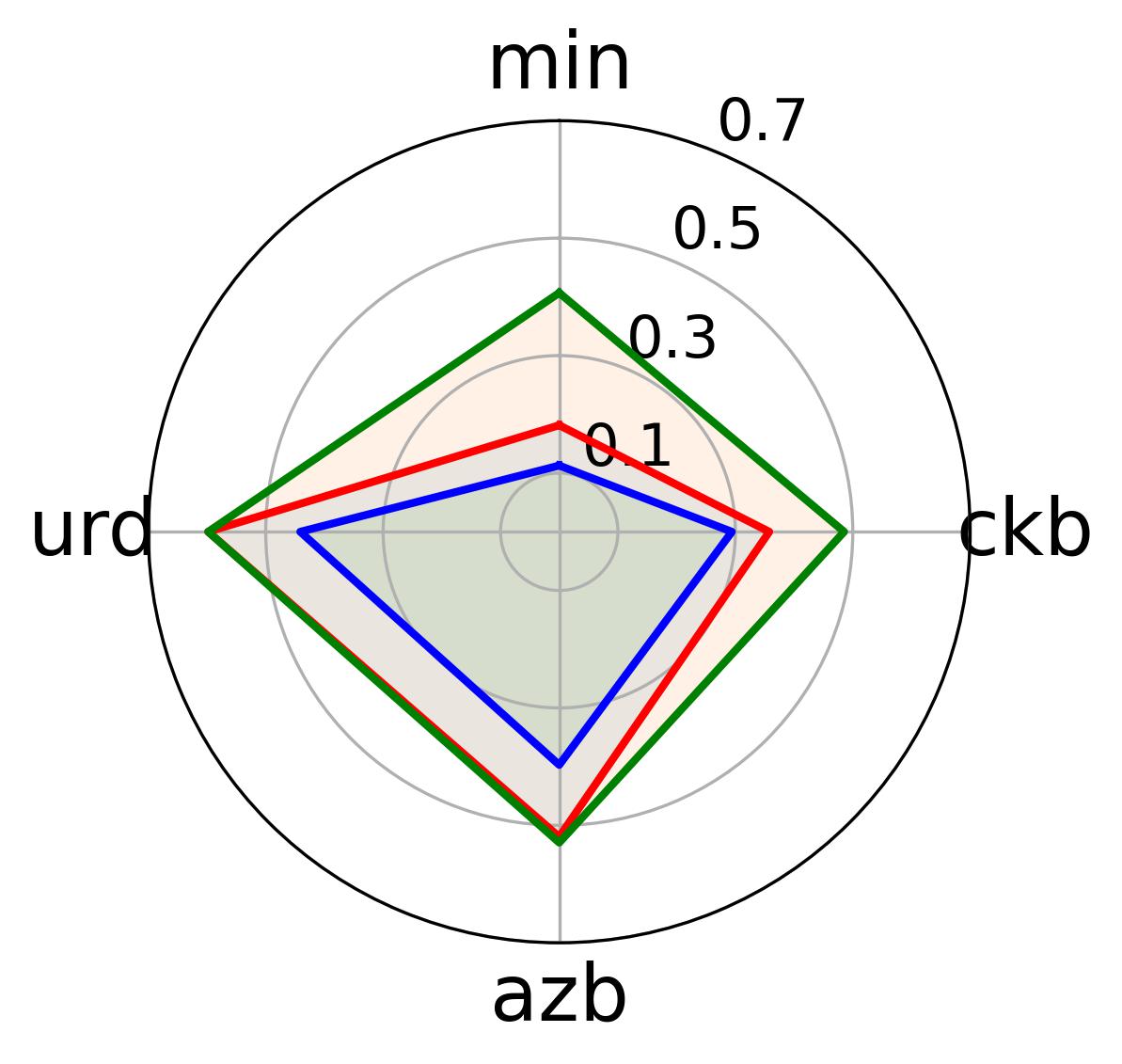}
\caption{Arabic}\label{fig:zero-arb}
\end{subfigure}
\begin{subfigure}{0.190\textwidth}
\includegraphics[width=1\linewidth]{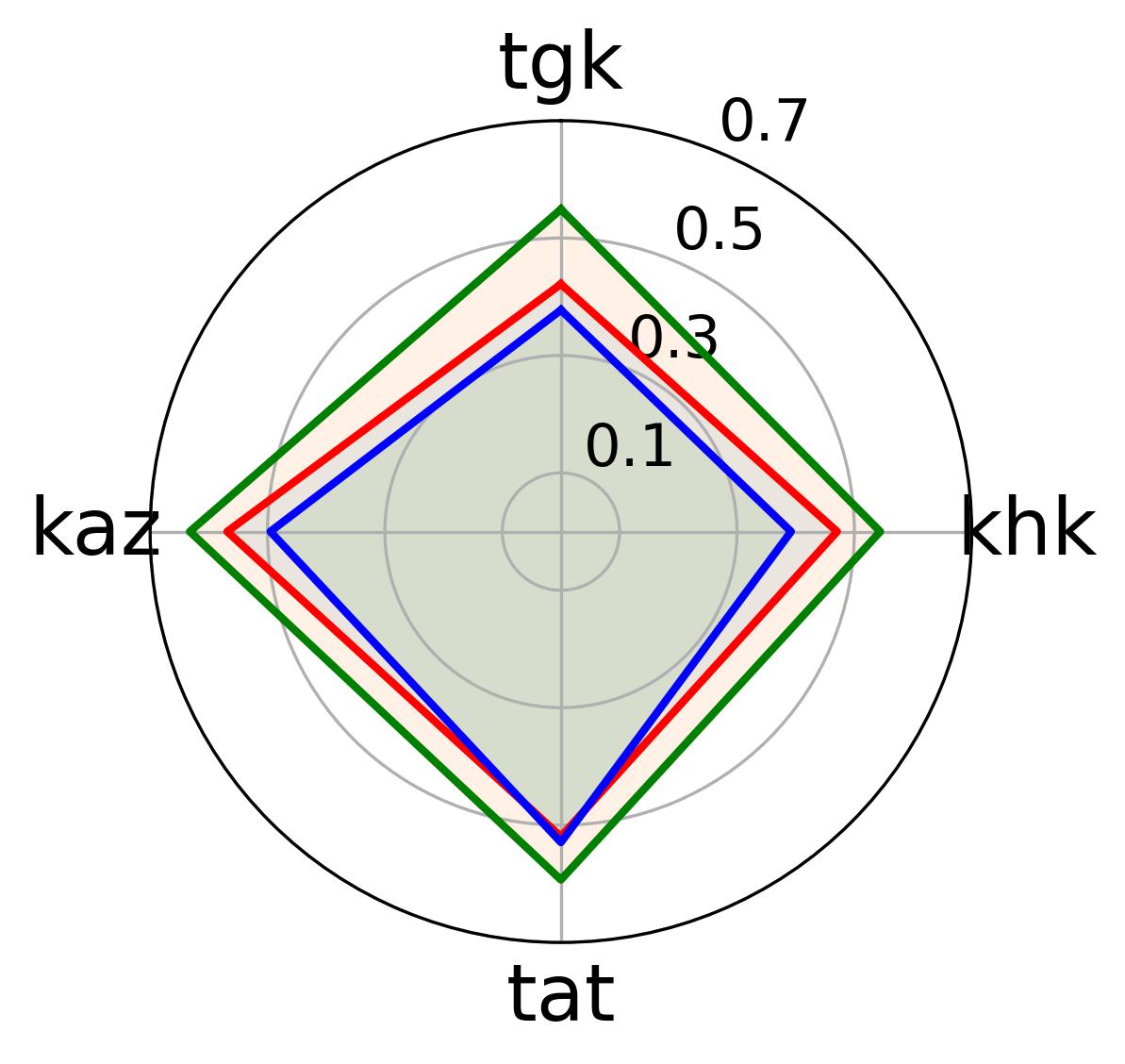}
\caption{Cyrillic}\label{fig:zero-cyl}
\end{subfigure}
\caption{Accuracy scores for DeepSeek in zero-shot ICL with sentence-level alignment when one unlabelled sentence is BM25-based (green) or randomly sampled (blue). Red denotes baseline zero-shot ICL.}
\label{fig:zero-result-align}
\end{figure}

\paragraph{Effectiveness of Semantic Similar Examples} 
Figure \ref{fig:zero-result-align} shows DeepSeek's performance when prompted with 
%We present the model performance when prompting DeepSeek with 
one unlabeled example in the target language alongside its English translation. Similar trends are observed for LLaMA-3.2 and Gemma-2, with detailed results in the Appendix \ref{app:results}. Although the topic label of the example is not included in the prompt, incorporating semantically similar texts in both target language and English significantly enhances performance for low-resource languages with low baseline zero-shot ICL performances, especially those using rarer scripts. However, this benefit diminishes as the baseline zero-shot ICL performance improves. For languages with baseline accuracy scores below 0.3, all three LLMs show an average improvement exceeding 0.22, with a peak gain of 0.36 for \texttt{sat} on LLaMA-3.2. In contrast, for languages with baseline scores above 0.3, the average improvement falls below 0.07. In some cases, LLaMA-3.2 and Gemma-2 exhibit minor performance declines, with the largest drops being 0.1 for \texttt{kaz} on both LLaMA-3.2 and Gemma-2 (baseline $=$ 0.59 and 0.61 respectively).
Furthermore, the models' performance often degrade when examples are randomly sampled from the training set, highlighting that the effectiveness of the alignment hinges on the semantic similarity between the input text and the example in target language. Moreover, improving semantic search for low-resource languages or even enhancing text pre-processing approaches (e.g., lemmatisation and stemming), beyond the simple whitespace-based tokenisation used here, has the potential to increase performance in this sentence-alignment setting.

\paragraph{Impact of the Number of Examples} We further analyse the performance when varying the number of unlabelled parallel examples provided in the prompt, from two to five examples. We find that increasing the number of randomly sampled parallel examples in the prompt could slightly improve the performance, although the gap between BM25 sampled examples is still notable. However, providing more semantic related examples does not consistently improve performance across all languages. We hypothesize that it may be influenced by the relative length of the target-language sequences compared to English.
%as ICL performance may degrade when relevant information (i.e., English translations) appears in the middle of long contexts \citep{liu-etal-2024-lost}. 
To test this, we compute the average \textit{tokenizer parity} (TP) scores \citep{petrov2023language} for each language in the training set (in Table \ref{tab:language full list}). Given a text in its target language and English translation, TP is defined as the ratio of the number of tokens in English to the number of tokens in the target language. A lower TP score indicates that the target language is tokenized into relatively longer sequences compared to English. We define a binary variable indicating whether adding more examples is beneficial: it is set to true if at least 3 out of the 4 multi-shot settings (2, 3, 4, and 5 examples) outperform the 1-shot setting. We then calculate the point-biserial correlation coefficient \citep{lev1949point} between the TP score and this binary indicator. The results show a statistically significant correlation, suggesting that languages with lower TP scores are less likely to benefit from additional parallel unlabelled examples.

\subsubsection{Word-Level Alignment} 

%\begin{figure}[h!]
%\centering
%\includegraphics[width=.35\textwidth]%{latex/image/dictionary_zeroICL.jpg}
%\caption{Performance on deepseek in zero-shot ICL setting with word-level alignment or translation in the prompt.}
%\label{fig:zeroICL-dict-results}
%\end{figure}

\begin{figure}[h!]
\centering
\begin{subfigure}{0.19\textwidth}
\includegraphics[width=1\linewidth]{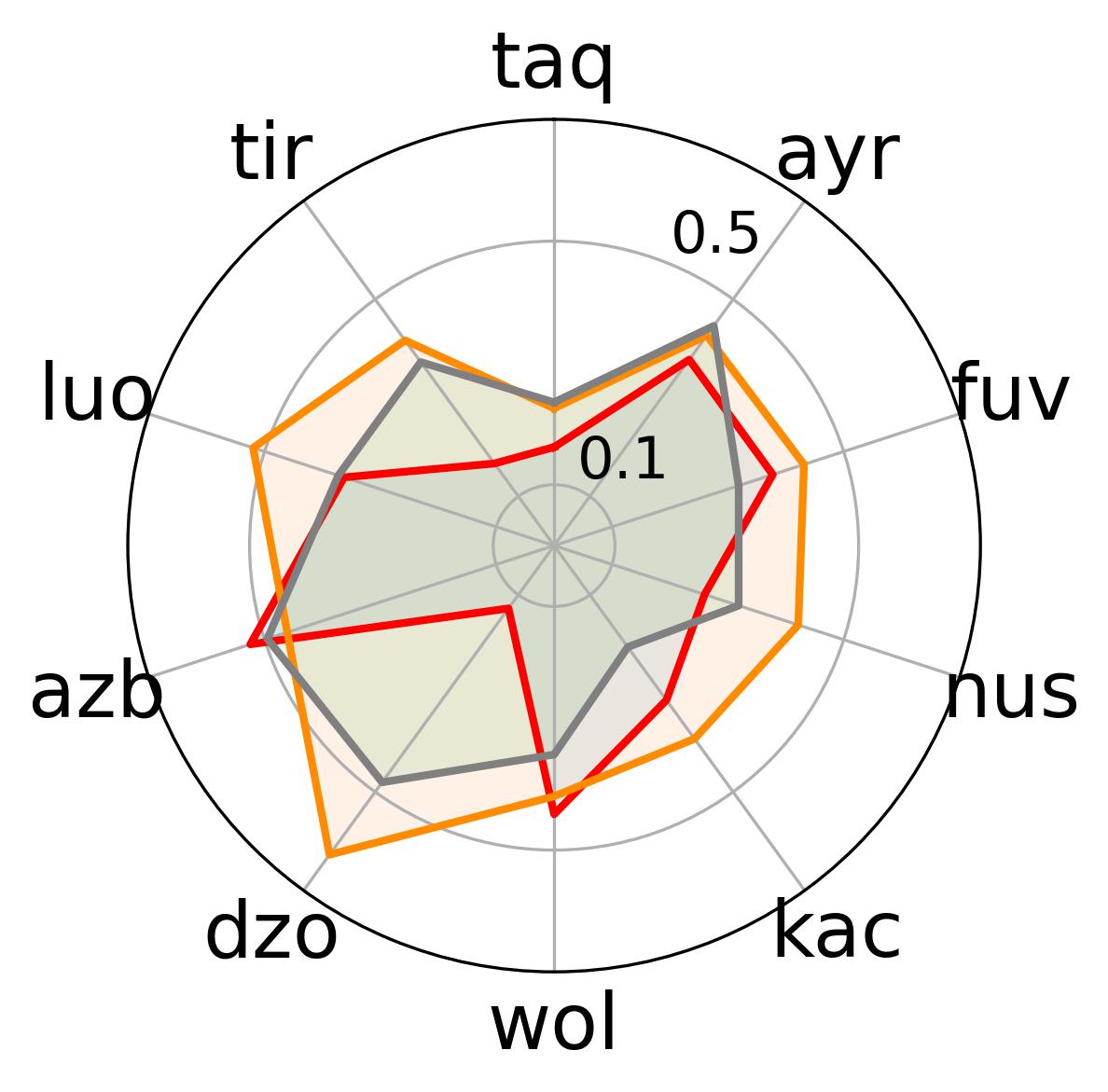}
\caption{chrf++ $\leq$ 0.5}\label{fig:dict_lowchrf}
\end{subfigure}
\begin{subfigure}{0.19\textwidth}
\includegraphics[width=1\linewidth]{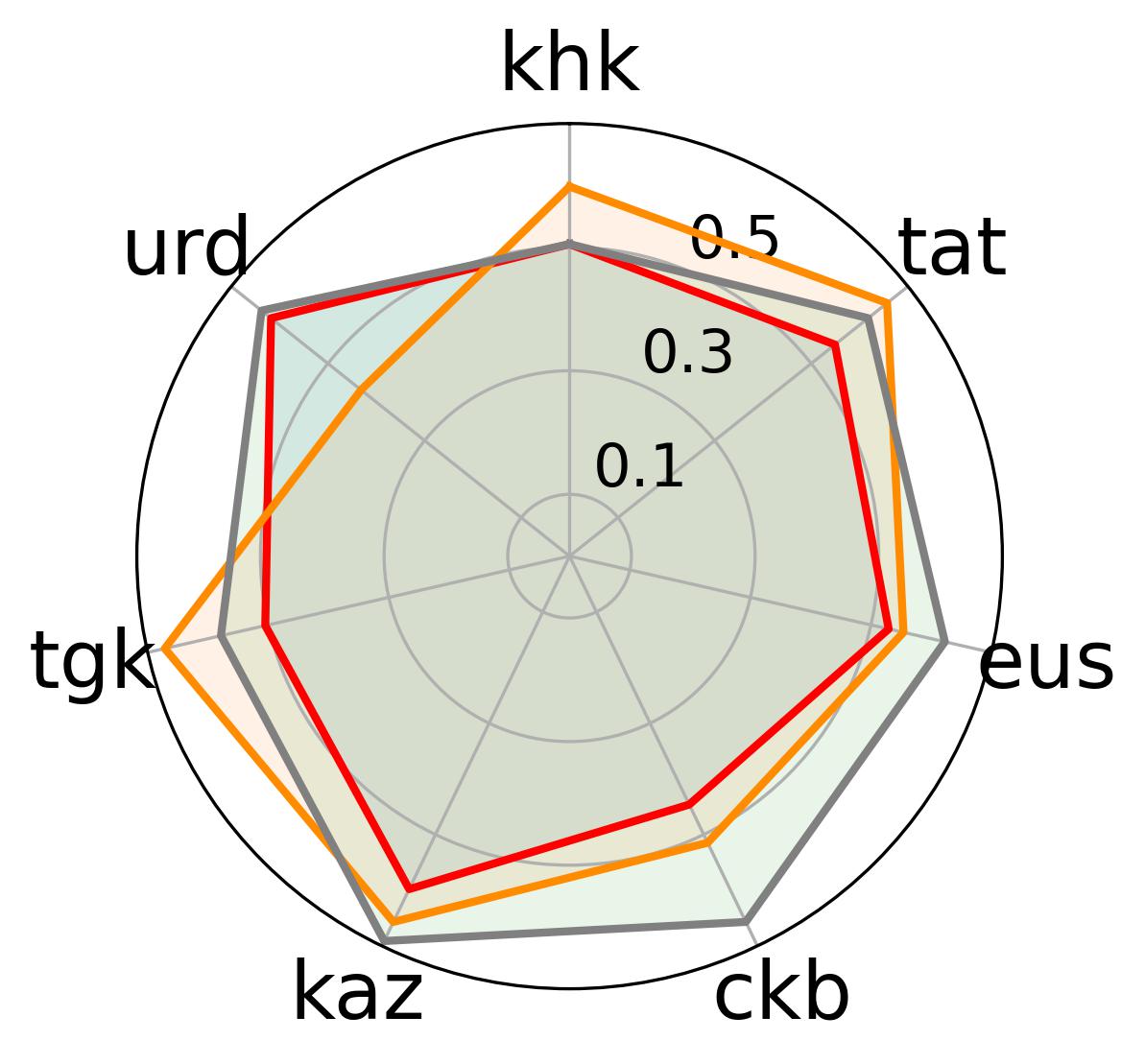}
\caption{chrf++ $>$ 0.5}\label{fig:dict_highchrf}
\end{subfigure}
\caption{Accuracy scores for LLaMA-3.2 over low-resource languages in zero-shot ICL with word-level alignment (gray) or word-level translation (orange) settings. Red denotes baseline zero-shot ICL.}
\label{fig:zero-word-align}
\end{figure}

The zero-shot ICL performance with word-level alignment or translation on LLaMA-3.2 is shown in Figure \ref{fig:zero-word-align}. The reported chrf++ score\footnote{\url{https://tinyurl.com/nllb200dense3bmetrics}} of each language on NLLB is used as an indicator of the quality of the dictionary. 
For DeepSeek and Gemma-2 (full results in Appendix \ref{app:results}), adopting word-level alignment or translation always improve the performance over the baseline. However, LLaMA-3.2's performance is highly influenced by the dictionary's quality. When the chrf++ score is lower than 0.5 (Figure \ref{fig:dict_lowchrf}), including the low-quality word-level alignment (gray lines) can harm the performance (e.g., \texttt{fuv}, \texttt{kac}, \texttt{wol}, and \texttt{azb}). In contrast, when chrf++ score is higher than 0.5 (Figure \ref{fig:dict_highchrf}), word-level alignment is always beneficial. For \texttt{nqo}, \texttt{sat}, and \texttt{min}, we train \textit{fast\_align} on the SIB-200 dataset to simulate a high-quality dictionary (see Section \ref{sec:learning}). LLMs' performance for these languages is always improved with word-level alignment (around 0.6 of accuracy improvement), highlighting the importance of the dictionary quality.

ICL with word-level translation significantly reduces the inference time than using word-level alignment by reducing input length.. 
However, its performance exhibits variable superiority across languages and LLMs. Specifically, it is consistently better than world-level alignment on LLaMA-3.2, while it is always worse than world-level or even baseline on DeepSeek and Gemma-2. %We hypothesise that the effectiveness of this approach is also impacted by the robustness of LLMs to noises in English, such as disordered words.

\subsection{Alignment in Few-Shot ICL} \label{sec:fewicl}

%\paragraph{Effectiveness of the Alignment} 
For languages with baseline accuracy lower than majority voting, including English translations in few-shot ICL often improves results across all the LLMs, when using more than one demonstration example. However, in 1-shot ICL, removing the English translation tends to yield better performance on DeepSeek and LLaMA-3.2, while Gemma-2 continues to benefit from them. For languages with strong zero-shot ICL performance, both DeepSeek and Gemma-2 benefit from the alignment, whereas LLaMA-3.2 performs best without them. 
Overall, unlike the consistent trends across LLMs in zero-shot ICL with alignment, we observe more variations how LLMs respond to aligned prompts in few-shot scenarios (full results in Appendix \ref{app:results}). 

%\paragraph{Prompt Sensitivity} 
Although not examined by prior work \citep{cahyawijaya-etal-2024-llms}, we find that model performance can be highly sensitive to the order of the task description and demonstration examples in the prompt, for certain languages, especially on DeepSeek. For example, when prompting Deepseek with 1-shot labelled \texttt{nqo} example with English translation, the accuracy jumps from 0.30 to 0.42 if the task description is moved to after the demonstration example. In most of the cases, prompting with demonstrations at the beginning lead to better performance for DeepSeek and Gemma-2, while LLaMA-3.2 slightly prefers task description at the beginning. Results presented for few-shot ICL are based on the optimal task description position selected on the validation set.

\subsection{Comparison across PEFT, Zero- and Few-Shot ICL}\label{sec:compareall}

Based on our analysis, zero-shot ICL with word-\footnote{Results based on \textit{fast\_align} are excluded, as it potentially leaks gold standard English translations into the prompt.} or sentence-level alignment and few-shot ICL with or without alignment can all achieve promising improvement over the baseline across languages and LLMs. Next, we discuss which approach is more effective from different aspects. We present the results in Figure \ref{fig:compare all} for LLaMA-3.2. DeepSeek and Gemma-2 show same trend (see Appendix \ref{app:results}).

\begin{figure}[t!]
\centering
\begin{subfigure}{0.1522\textwidth}
\includegraphics[width=1\linewidth]{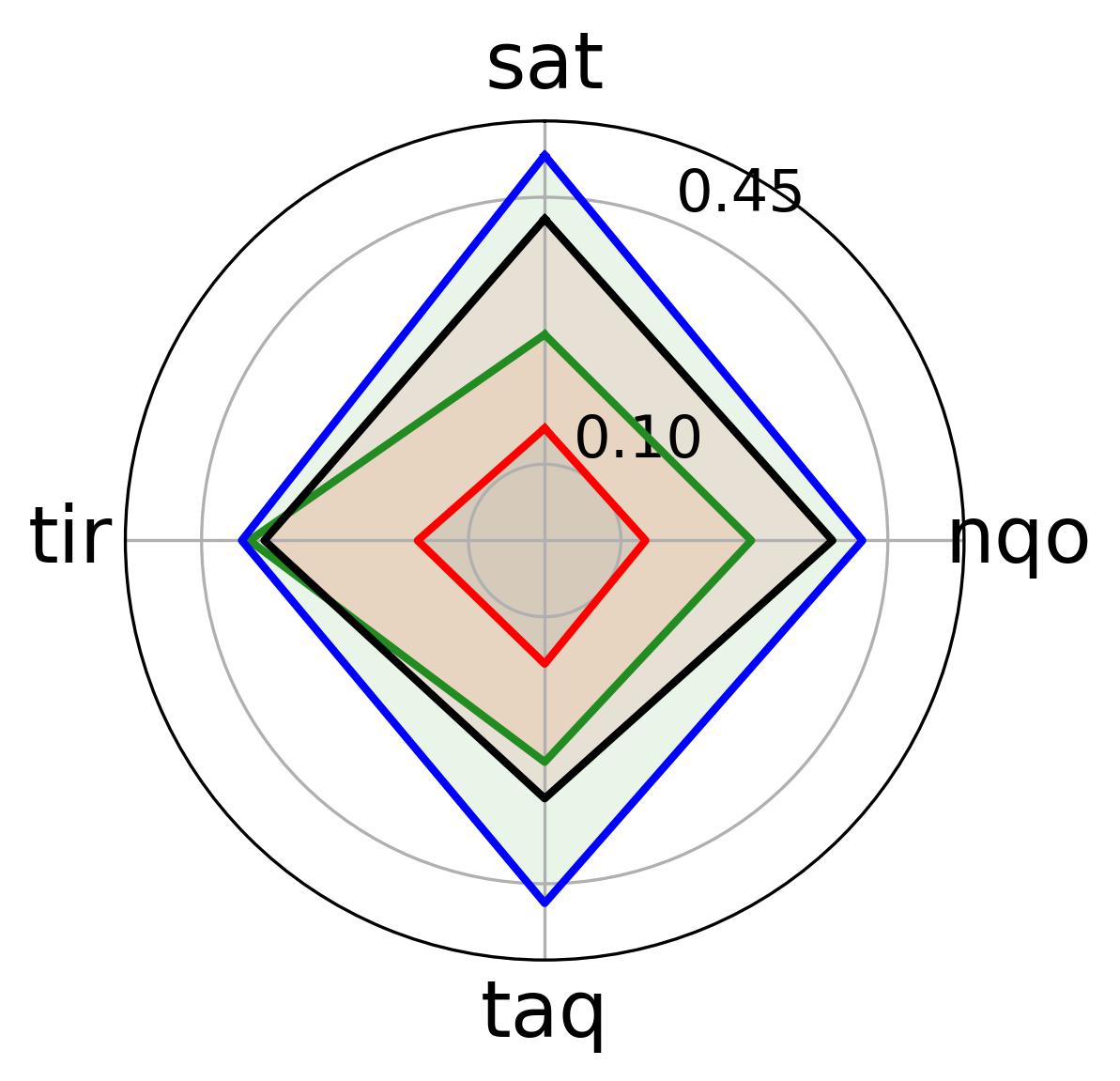}
\caption{extremely low}\label{fig:compare all-rare}
\end{subfigure}
\begin{subfigure}{0.1522\textwidth}
\includegraphics[width=1\linewidth]{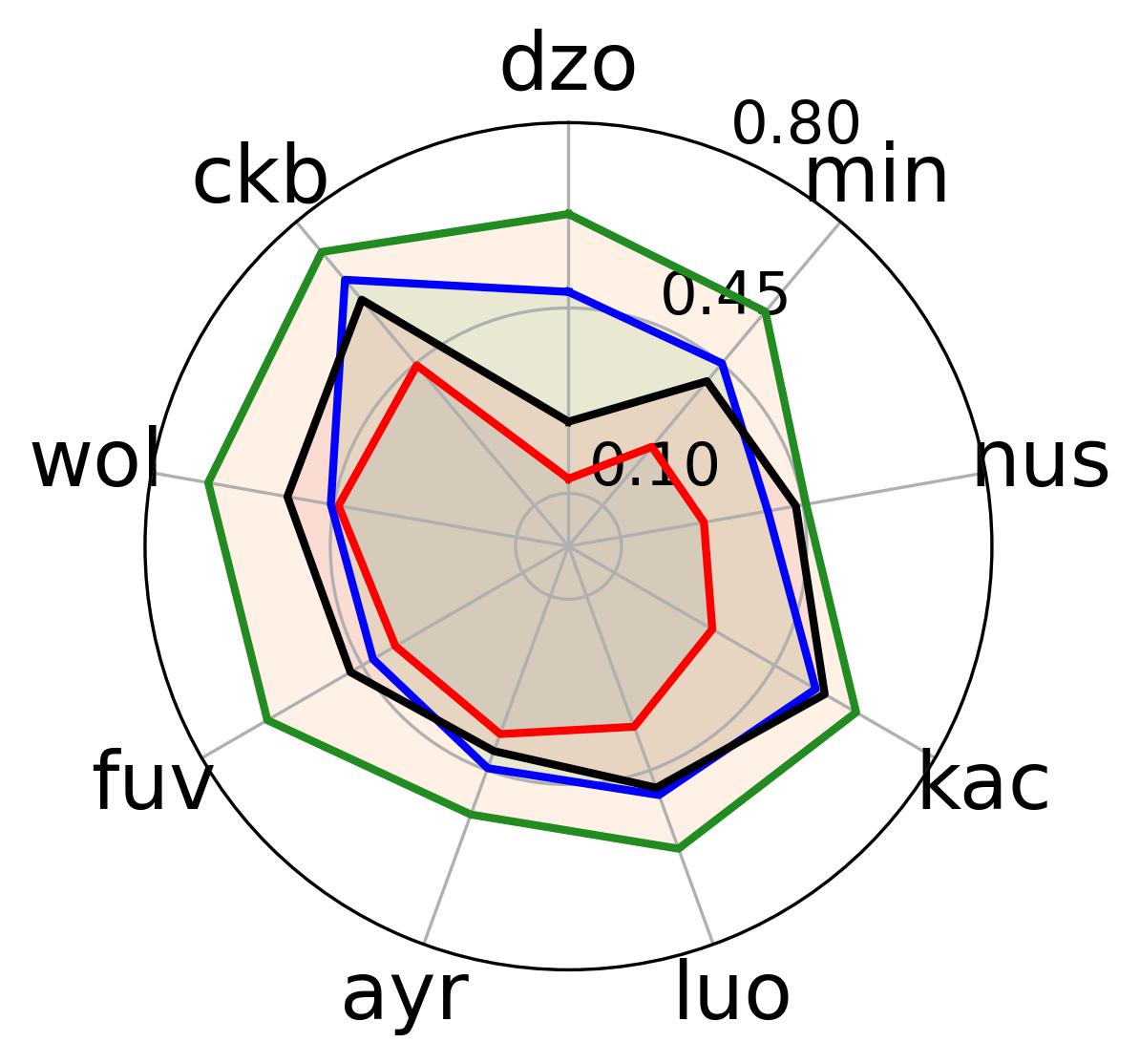}
\caption{low (acc<0.45)}\label{fig:compare all-baseline<0.45}
\end{subfigure}
\begin{subfigure}{0.1522\textwidth}
\includegraphics[width=1\linewidth]{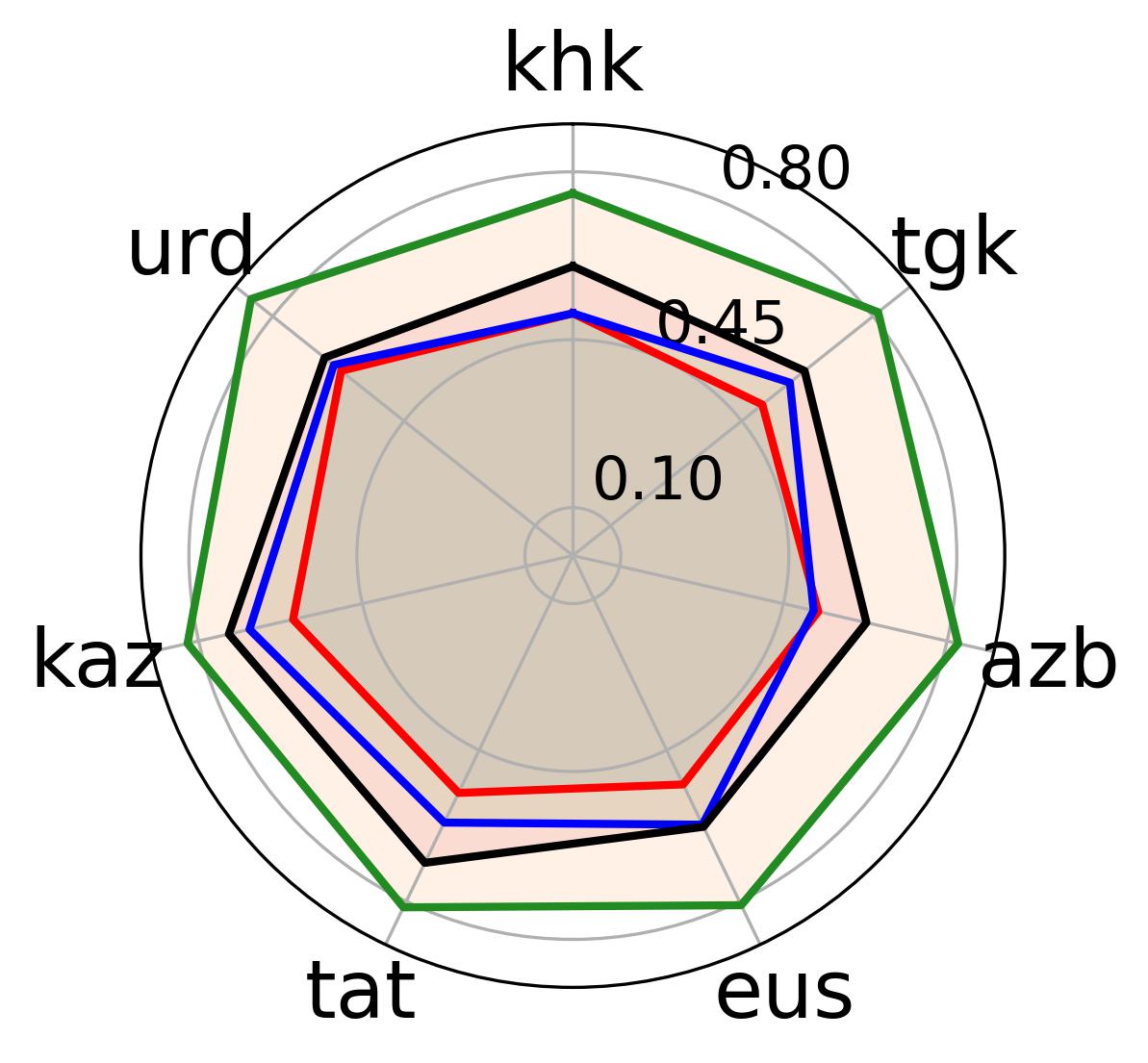}
\caption{low (acc>0.45)}\label{fig:compare all-baseline>0.45}
\end{subfigure}
\caption{Accuracy comparison among baseline (\textcolor{red}{red}), PEFT (\textcolor{ForestGreen}{green}), best zero-shot ICL (\textcolor{blue}{blue}) and best few-shot ICL (\textbf{black}) on LLaMA-3.2. Languages are categorised into: 
(a) Both language and script are severely under-represented (names in red in Figure \ref{fig:ip-ft-llama32}, baseline accuracy < 0.2): \textbf{zero-shot $>$ few-shot $>$ PEFT}; (b) Better represented, but baseline < 0.45: \textbf{PEFT $>$ zero-shot $\geq$ few-shot}; 
(c) Baseline > 0.45: \textbf{PEFT $>$ few-shot $>$ zero-shot}.}
\label{fig:compare all}
\end{figure}

\paragraph{Fine-Tuning vs. ICL}

When the low-resource language is extremely under-represented in both tokeniser and model (Figure \ref{fig:compare all-rare}), fine-tuning LLMs or even PLMs\footnote{Based on results for XLM-R (large) from \citet{adelani-etal-2024-sib}, see Table \ref{tab:PEFT full results}.} yields minimal improvement, while zero-shot ICL with either sentence- or word-level alignment offers significant improvements (Figure \ref{fig:ft-deepseek} vs. \ref{fig:zero-rare}). \citet{adelani-etal-2024-sib} extended the vocabulary of XLM-R for \texttt{nqo} with continue-pretraining, leading to fine-tuning accuracy on SIB-200 test set rising 0.17 points. However, it is still lower than the performances of LLMs in zero-shot ICL with alignment, which are above 0.41. 

For the remaining low-resource languages (Figure \ref{fig:compare all-baseline<0.45} and \ref{fig:compare all-baseline>0.45}), fine-tuning normally has better performance than ICL. Overall, the average difference between PEFT and the best ICL approach on DeepSeek, LLaMA-3.2 and Gemma-2 is 0.13, 0.13, and 0.08, respectively. %, with the highest as 0.21 (fuv on DeepSeek) and the lowest as 0.02 (nus on LLaMA-3.2).

\paragraph{Zero-Shot vs. Few-Shot} 

We observe that for languages with low baseline zero-shot ICL performance (i.e., accuracy < 0.45 on all LLMs), in most cases (at least more than 50\%), zero-shot ICL with word/sentence-level alignment leads to better performance than few-shot ICL regardless if the alignment is provided or not (Figure \ref{fig:compare all-baseline<0.45}, comparing black and blue lines). When baseline performance is higher than 0.45 (Figure \ref{fig:compare all-baseline>0.45}), few-shot ICL always provides best results. In our study, these languages are \texttt{khk}, \texttt{tgk}, \texttt{azb}, \texttt{eus}, \texttt{tat}, \texttt{kaz}, and \texttt{urd}, consistent across all the LLMs. Overall, if the LLM significantly lacks capability on the target language, providing label in ICL might be useless.

\section{Discussion}

\subsection{NLU Tasks beyond Topic Classification}

We test our findings on BELEBELE, a reading comprehension parallel multilingual dataset, covering 11 out of the 20 languages studied here. It contains questions with four multiple-choice answers linked to a passage. \texttt{tir} is available among the five languages with rare scripts. We split the data into training, validation and test, following SIB-200, to enable a consistent comparison (See Appendix \ref{app:results}). We conduct experiments only on LLaMA-3.2, due to its long context length and computational constraints. 
%We focus on the approaches that show promising results on SIB-200: PEFT, zero-shot ICL with word or sentence alignment, baseline few-shot ICL, and few-shot ICL with alignment. 
We retrieve the passage from the training data with BM25 as example and provide its English translation as passage alignment. We adopt accuracy for evaluation \citep{bandarkar-etal-2024-belebele}.

Most results align with our observations on SIB-200 dataset. Specifically, PEFT still shows no improvement for \texttt{tir}, while being more effective for other languages. Zero-shot ICL with passage alignment could still improve over the baseline zero-shot ICL in most cases, especially for the languages with lower baseline performance (e.g., accuracy < 0.35). However, as the task is more challenging, the level of improvement is not as notable as on topic classification. The model also potentially requires more unlabelled parallel data for consistent improvements across all languages. Similar to topic classification, in most cases, zero-shot ICL with passage alignment surpasses few-shot ICL when baseline performance is lower than 0.5. 

Conversely, word-level alignment is not effective on reading comprehension, which may be explained by the quality of the dictionary created. Unlike topic classification, whose prediction can be made based on one or two topic-related words, reading comprehension relies less on such cue words, requiring a higher quality of word translations. %The large number of incorrect translations just serves as noises in ICL.

\subsection{Suggestions to Practitioners} \label{sec:suggestion}

In practice, performance is not the only consideration. Investment in data and computational resources needs to be carefully considered, especially for low-resource languages. Aiming to adapt an LLM to a low-resource language for a downstream task, which approach should be prioritized, and what types of data should be created? 

%We give following suggestions for practitioners aiming to adapt an LLM to a low-resource language for a downstream task.

\paragraph{For low-resource languages that are extremely under-represented in both tokeniser and model} (e.g., \texttt{nqo}) fine-tuning is not effective and zero-shot ICL with alignment shows promising improvements. We suggest \textit{prioritizing investment in human translation to create a small-scale in-domain parallel data for zero-shot ICL with alignment}.

\paragraph{For low-resource languages where LLMs demonstrate limited capability} few-shot ICL might lead to better performance than zero-shot ICL with alignment. However, in most cases, these gains are modest and may even come at the risk of performance degradation. With fine-tuning LLM/PLM being effective and with acceptable zero-shot ICL performance for these languages, \textit{decisions should be made by comparing the financial costs between human translation (for zero-shot ICL) and human annotation (for fine-tuning)}.

\paragraph{For low-resource languages where LLMs demonstrate a certain level of capability} few-shot leads to better performance than zero-shot ICL with alignment. Human annotation tends to be required for a notable improvement for these languages. Practitioners should \textit{consider the trade-off between the amount of data to annotate (effective fine-tuning may require more data than few-shot ICL) and the computational costs (LLM inference is more expensive than fine-tuning PLMs)}.

\section{Conclusion}

This work provides a systematic analysis on whether ICL can enable LLMs to effectively support extremely low-resource languages on downstream tasks. As some of the key findings that contrast to prior work, we reveal the limitation of fine-tuning when languages and their scripts are both highly under-represented. In such cases, zero-shot ICL augmented with word- or sentence-level alignment yields promising results. Meanwhile, few-shot ICL or PEFT tends to perform better for languages relatively better represented during pretraining. Our study highlights the importance of language and script coverage in LLMs, and the strong potential of ICL for language adaptation.

%\clearpage

\section*{Limitations}

Although we conducted more than 450 experiments, our study did not include other popular LLMs, such as Mistral and Qwen. Due to our computational constraints and consideration of fair comparison with PEFT on same LLM size, we did not experiment with LLMs with large sizes, such as Gemma-2 (9b) \footnote{\url{https://huggingface.co/google/gemma-2-9b}} or LLaMA-3.3 (70b) \footnote{\url{https://huggingface.co/meta-llama/Llama-3.3-70B-Instruct}}. Future work could explore whether a larger LLM could enable even more improvement in ICL with word- or sentence level language alignment. 

Due to very limited datasets with parallel data available for these extremely low-resource languages, we covered topic classification and reading comprehension in this study. On reading comprehension, we were only able to experiment with 11 of the 20 target languages. Both SIB-200 and BELEBELE are constructed based on Flores-200 dataset \citep{costa2022no}. With the increase of language coverage for NLP tasks in the future, our findings could be tested on other tasks (e.g., common-sense reasoning or summarization.) and other domains (e.g., medical, social media).

As a lack of native speakers and reliable gold-standard word translations, we were not able to accurately access the quality of the dictionary that we created using NLLB translator or fastalign. Our study does not show promising results when using the created dictionary to assist reading comprehension in ICL. However, as discussed in the main content, the results might be improved with a better dictionary. But the effectiveness on SIB-200 and ineffectiveness on BELEBELE imply that for challenging tasks with long input length, word translation quality is more important than for tasks with shorter input. Also, we directly translate words into English to simulate a dictionary following prior work. However, in practice, using a real-world dictionary raises additional challenges such as handling lexical ambiguity and polysemy, which may also impact the performance of word-level alignment in zero-shot ICL. 

As very limited parallel corpus available for our target languages, we did not systematically analyse how the ICL performance would be impacted if the included unlabelled parallel text is out-of-domain (e.g., from another dataset). However, since randomly sampled examples from the training data already poses risk of performance degradation, we hypothesise that zero-shot sentence-level alignment with out-of-domain examples might demonstrate limited benefit.

\section*{Acknowledgments}

This work is supported by the UK’s innovation agency (InnovateUK) grant number 10039039 (approved under the Horizon Europe Programme as VIGILANT, EU grant agreement number 101073921) (https://www.vigilantproject.eu).

\bibliography{custom,anthology}

\appendix

\section{Languages and Datasets Information}
\label{sec:appendix language info}

\paragraph{Languages} The languages we experiment with are persented in Table \ref{tab:language full list}, along with their IP and TP scores across LLMs. The chrf++ score from English to target language with NLLB translator, which we used to create the dictionary, is also included.

\paragraph{Datasets} SIB-200 dataset is constructed based on the Flores-200 dataset. The data is categorised into seven topic classes: science/technology, travel, politics, sports, health, entertainment, and geography. The official training, validation and test set contain 701, 99 and 204 data points, respectively.

BELEBELE dataset is also derived from the Flores-200 dataset. It contains a passage, a question linked to the paragraph, and four choices. Following SIB-200, we split the dataset into training, validation and test set with 600, 93, and 207 data samples. No overlapping between the passages in the training/validation and test set. We preliminarily test two different random train/validation/test splits on Tigrinya and find the results are consistent.

\begin{table*}[h!]
\centering
\scalebox{0.6}{
\begin{tabular}{llll|ccc|ccc|cc}
\hline
\textbf{Language Code} & \textbf{Language} & \textbf{Script} & \textbf{Family} & \multicolumn{3}{c|}{\textbf{Information Parity}} & \multicolumn{3}{c}{\textbf{Tokenizer Parity}} & \multicolumn{2}{|c}{\textbf{NLLB}}\\
& & & & DeepSeek & LLaMA & Gemma & DeepSeek & LLaMA & Gemma & eng-X & X-eng\\
\hline
nqo\_Nkoo & Nko & NKo & Manding & 0.16 & 0.15 & 0.16 & 0.10 & 0.10 & 0.17 & - & -\\
sat\_Olck & Santali & Ol Chiki  & Austroasiatic & 0.17 & 0.27 & 0.28 & 0.08 & 0.08 & 0.20 & 28.4 & 39.9\\
taq\_Tfng & Tamasheq & Tifinagh & Afro-Asiatic & 0.18 & 0.22 & 0.19 & 0.13 & 0.11 & 0.21 & 18.8 & 26.2\\
tir\_Ethi & Tigrinya & Ge'ez & Afro-Asiatic & 0.20 & 0.26 & 0.25 & 0.13 & 0.14 & 0.31 & 24.8 & 49\\
dzo\_Tibt & Dzongkha	& Tibetan & Sino-Tibetan & 0.20 & 0.25 & 0.22 & 0.08 & 0.09 & 0.26 & 32.6 & 40.1\\
nus\_Latn & Nuer & Latin & Nilotic &	0.21 & 0.22 & 0.22 & 0.31 & 0.27 & 0.37 & 28.9 & 38.2\\
min\_Arab & Minangkabau & Arabic & Austronesian & 0.22 & 0.23 & 0.22 & 0.24 & 0.37 & 0.43 & - & -\\
tgk\_Cyrl & Tajik & Cyrillic & Indo-European & 0.23 & 0.28 & 0.32 & 0.36 & 0.36 & 0.42 & 49.8 & 59.5\\
ayr\_Latn & Central Aymara & Latin &  Aymaran & 0.24 & 0.26 & 0.25 & 0.49 & 0.49 & 0.54 & 29.6 & 28.7\\
kac\_Latn & Jingpho & Latin &  Sino-Tibetan & 0.24 & 0.25 & 0.25 & 0.45 & 0.46 & 0.51 & 38 & 39.3\\
wol\_Latn & Wolof & Latin & Atlantic-Congo & 0.25 & 0.28 & 0.27 & 0.53 & 0.56 & 0.61 & 28.1 & 39.8\\
azb\_Arab & South Azerbaijani & Arabic & Turkic & 0.25 & 0.29 & 0.31 & 0.28 & 0.55 & 0.58 & 23.8 & 43.6\\
tat\_Cyrl & Tatar & Cyrillic & Turkic & 0.25 & 0.32 & 0.37 & 0.34 & 0.34 & 0.47 & 48.7 & 56.7\\
luo\_Latn & Luo & Latin & Nilotic & 0.26 & 0.28 & 0.27 & 0.55 & 0.57 & 0.61 & 39 & 45.8\\
fuv\_Latn & Nigerian Fulfulde & Latin & Atlantic-Congo & 0.28 & 0.31 & 0.30 & 0.58& 0.59 & 0.65 & 23.2 & 32.4\\
ckb\_Arab & Central Kurdish & Arabic &  Indo-European & 0.30 & 0.32 & 0.35 & 0.21 & 0.26 & 0.36 & 45.2 & 58.5\\
khk\_Cyrl & Halh Mongolian & Cyrillic & Mongolic-Khitan & 0.32 & 0.33 & 0.38 & 0.33 & 0.33 & 0.40 & 42 & 52.6\\
eus\_Latn & Basque & Latin & Basque & 0.32 & 0.44 & 0.45 & 0.54 & 0.56 & 0.62 & 48.5 & 57.5\\
kaz\_Cyrl & Kazakh & Cyrillic & Turkic & 0.32 & 0.36 & 0.46 & 0.32 & 0.35 & 0.45 & 50.7 & 59.4\\
urd\_Arab & Urdu & Arabic & Indo-European & 0.36 & 0.52 & 0.49 & 0.22 & 0.34 & 0.54 & 48.3 & 61.7\\
\hline
\end{tabular}
}
\caption{Full list of the 20 languages we experiment with in this study from the SIB-200 dataset, along with their information parity and tokenizer parity scores on DeepSeek, LLaMA-3.2 and Gemma-2. The reported chrf++ scores from two directions (eng-X: English to target language; X-eng: target language to English) with NLLB-200 translator (3.3B variant) is also included. Language code represents language (ISO 639-3)\_script (ISO 15924).}
\label{tab:language full list}
\end{table*}

\section{Implementation Details} \label{app:imp}

\paragraph{Prompt} For BELEBELE, we adopt the same prompt used by the authors for baseline zero-shot ICL\footnote{\url{https://github.com/facebookresearch/belebele/blob/main/sample_zero_shot_instructions.md}}. As for SIB-200, we use the following prompt for baseline zero-shot ICL: "\textit{What is the topic discussed in the following \{language name\} text? There are seven options: "science/technology", "travel", "politics", "sports", "health", "entertainment", and "geography". Now complete the following example without explanations. Text: \{text\}. Topic option is:}", as we found it performing better on the validation set than the one used by the original authors of SIB-200. Additionally, we observed that explicitly indicating the language of the input text had no impact on performance. For extremely low-resource languages such as Nko and Santali, LLaMA-3.2 and Gemma-2 refuse to perform the task if prompted with "....\textit{complete the following example}", stating that they do not recognize the input language, no matter whether the name of the language is explicitly given in the prompt or not. However, they would produce a prediction when prompted with "\textit{complete the following example without explanations}". For sentence-level alignment, the LLMs are instructed as "\textit{Use the following pairs of \{language name\} texts and their English translations to help you understand \{language name\}.\{alignment example\}. Now based on your understanding, answer the question below without explanation.}". For word-level alignment, we instruct LLMs with "\textit{Please use the provided English translation of each word to help you understand the \{language name\} text}.". Experiments are conduct on NVIDIA A100-PCIE-40GB.

\paragraph{Dictionary} For each data sample in the test set, we extract words in target language based on white-space splitting only. Then we use NLLB-200 translator (3.3B)\footnote{\url{https://huggingface.co/facebook/nllb-200-3.3B}} to translate each word into English. For the three languages that are not supported by NLLB, we train the word alignment tool \textit{fast\_align} \citep{dyer-etal-2013-simple} with SIB-200 training data and then align the English words and target-language words in the test set. We use the default training and alignment settings in \textit{fast\_align}\footnote{\url{https://github.com/clab/fast_align}}.

\paragraph{IA3} The rescale vectors are learnt for key, value of the attention modules and feed-forward network in each layer. We use batch size as 4, and early stopping strategy based on validation loss, with max training epochs as 10. We use AdamW \citep{loshchilov2018decoupled} for optimization. We perform hyper-parameter search on learning rate of \{1e-3, 5e-3, 8e-3, 1e-2\}. The optimal learning rate on validation set is 8e-3. For extremely low-resource languages where IA3 show limited improvement, we also search learning rate from \{1e-4, 3e-3, 7e-3\}. We run experiments 3 times and report the average performance. Experiments are conducted on NVIDIA GH200 480GB.

\section{Full Results} \label{app:results}

\paragraph{BELEBELE} The results on LLaMA-3.2 are in Table \ref{tab:belebele full results}.

\begin{table*}[h!]
\centering
\scalebox{0.7}{
\begin{tabular}{lc|ccccc}
\hline
\textbf{Language Code} & \textbf{Baseline Zero-Shot} & \textbf{PEFT} & \textbf{Zero-Shot with Align} & \textbf{Baseline Few-Shot} & \textbf{Few-Shot with Align}\\
\hline
kac\_Latn&0.261&0.353(1)&0.329(2)&0.285&0.280\\
wol\_Latn&0.261&0.361(1)&0.319(2)&0.256&0.261\\
fuv\_Latn&0.266(2)&0.309(1)&0.256&0.227&0.237\\
tir\_Ethi&0.275&0.271&0.300(1)&0.227&0.280(2)\\
luo\_Latn&0.295&0.314(2)&0.280&0.319(1)&0.275\\
ckb\_Arab&0.333&0.440(1)&0.372(2)&0.280&0.324\\
tgk\_Cyrl&0.357&0.391(1)&0.386(2)&0.338&0.343\\
kaz\_Cyrl&0.362(2)&0.464(1)&0.348&0.275&0.251\\
khk\_Cyrl&0.372(2)&0.472(1)&0.290&0.266&0.300\\
eus\_Latn&0.415&0.623(1)&0.420(2)&0.367&0.338\\
urd\_Arab&0.517&0.638(1)&0.459&0.546(2)&0.473\\
\hline
\end{tabular}
}
\caption{The accuracy scores on the BELEBELE test set with LLaMA-3.2: baseline ICL (zero-shot), PEFT, zero-shot with alignment (3 parallel examples retrieved with BM25), 3-shot baseline ICL, and 3-shot ICL with alignment. Differences between baseline zero-shot ICL is statistical significant (paired chi-squared test). The number in parentheses denotes the rank of the performance on the target language.}
\label{tab:belebele full results}
\end{table*}

\paragraph{SIB-200} The results of baseline zero-shot ICL, PEFT, along with fine-tuning multilingual PLM (from \citet{adelani-etal-2024-sib}) are presented in Table \ref{tab:PEFT full results}. 

\begin{table*}[ht!]
\centering
\scalebox{0.9}{
\begin{tabular}{l|ccc|ccc|c}
\hline
\textbf{Language Code} & \multicolumn{3}{c|}{\textbf{Baseline Zero-Shot ICL}} & \multicolumn{3}{c|}{\textbf{PEFT}} & \textbf{XLM-R}\\
& DeepSeek & LLaMA & Gemma & DeepSeek & LLaMA & Gemma \\
\hline
taq\_Tfng & 0.118 & 0.162 & 0.147 & 0.309 & 0.290 & 0.461 & 0.269 \\
dzo\_Tibt & 0.128 & 0.127 & 0.132 & 0.495 & 0.627 & 0.676 & 0.242 \\
nqo\_Nkoo & 0.137 & 0.132 & 0.127 & 0.323 & 0.271 & 0.245 & 0.232 \\
sat\_Olck & 0.172 & 0.147 & 0.333 & 0.240 & 0.270 & 0.608 & 0.245 \\
tir\_Ethi & 0.186 & 0.167 & 0.363 & 0.456 & 0.387 & 0.632 & 0.677 \\
min\_Arab & 0.181 & 0.245 & 0.260 & 0.583 & 0.578 & 0.520 & 0.381 \\
nus\_Latn & 0.250 & 0.260 & 0.255 & 0.569 & 0.456 & 0.485 & 0.439 \\
ayr\_Latn & 0.260 & 0.377 & 0.333 & 0.637 & 0.539 & 0.559 & 0.525 \\
kac\_Latn & 0.265 & 0.314 & 0.319 & 0.672 & 0.627 & 0.574 & 0.627 \\
luo\_Latn & 0.289 & 0.363 & 0.382 & 0.652 & 0.608 & 0.623 & 0.600 \\
fuv\_Latn & 0.304 & 0.378 & 0.382 & 0.681 & 0.657 & 0.554 & 0.630 \\
ckb\_Arab & 0.358 & 0.446 & 0.446 & 0.603 & 0.725 & 0.716 & 0.501 \\
wol\_Latn & 0.387 & 0.441 & 0.436 & 0.657 & 0.691 & 0.632 & 0.601 \\
tgk\_Cyrl & 0.422 & 0.505 & 0.485 & 0.696 & 0.814 & 0.716 & 0.598 \\
khk\_Cyrl & 0.471 & 0.505 & 0.490 & 0.681 & 0.755 & 0.691 & 0.885 \\
eus\_Latn & 0.490 & 0.529 & 0.588 & 0.750 & 0.809 & 0.804 & 0.892 \\
azb\_Arab & 0.520 & 0.525 & 0.539 & 0.721 & 0.824 & 0.789 & 0.829 \\
tat\_Cyrl & 0.520 & 0.549 & 0.583 & 0.706 & 0.814 & 0.799 & 0.819 \\
kaz\_Cyrl & 0.569 & 0.598 & 0.618 & 0.765 & 0.824 & 0.877 & 0.914 \\
urd\_Arab & 0.598 & 0.618 & 0.608 & 0.662 & 0.858 & 0.848 & 0.876 \\
\hline
eng\_Latn & 0.828 & 0.770 & 0.647 & 0.926 & 0.926 & 0.931 & 0.921 \\
\hline
\end{tabular}
}
\caption{Baseline zero-shot ICL and PEFT performance over SIB-200 on DeepSeek, LLaMA-3.2 and Gemma-2. Differences between baseline zero-shot ICL is statistical significant (paired chi-squared test). Performance of fine-tuning XLM-R(\textit{large}) is adopted from \citet{adelani-etal-2024-sib}.}
\label{tab:PEFT full results}
\end{table*}

The results of zero-shot ICL with word-level, word translation and sentence-level alignment and few-shot ICL are presented in Table \ref{tab:PEFT zero-shot results}. 

\begin{table*}[h!]
\centering
\scalebox{0.7}{
\begin{tabular}{ll|cccc|cc}
\hline
\textbf{Model} & \textbf{Language Code} & \multicolumn{4}{c|}{\textbf{Zero-Shot}}&\multicolumn{2}{c}{\textbf{Few-Shot}}\\
& & \textbf{sentence(BM25)} & \textbf{sentence(random)} & \textbf{word} & \textbf{word translation}&\textbf{without align}&\textbf{with align}\\
\hline
DeepSeek&taq\_Tfng&0.466&0.098&0.265&0.221&0.338&0.328\\
&dzo\_Tibt&0.279&0.176&0.623&0.676&0.225&0.181\\
&nqo\_Nkoo&0.436&0.132&0.617&0.681&0.451&0.417\\
&sat\_Olck&0.456&0.147&0.632&0.563&0.358&0.284\\
&min\_Arab&0.407&0.113&0.621&0.627&0.377&0.368\\
&tir\_Ethi&0.392&0.196&0.368&0.397&0.387&0.343\\
&nus\_Latn&0.368&0.186&0.412&0.319&0.373&0.387\\
&ayr\_Latn&0.328&0.265&0.461&0.422&0.353&0.358\\
&kac\_Latn&0.574&0.279&0.431&0.260&0.451&0.431\\
&luo\_Latn&0.539&0.304&0.500&0.490&0.515&0.456\\
&fuv\_Latn&0.441&0.304&0.431&0.328&0.441&0.475\\
&ckb\_Arab&0.485&0.294&0.505&0.627&0.422&0.417\\
&wol\_Latn&0.505&0.353&0.505&0.392&0.505&0.534\\
&tgk\_Cyrl&0.549&0.377&0.578&0.588&0.529&0.608\\
&khk\_Cyrl&0.544&0.392&0.544&0.539&0.471&0.525\\
&eus\_Latn&0.495&0.407&0.613&0.618&0.574&0.554\\
&azb\_Arab&0.529&0.397&0.480&0.466&0.559&0.554\\
&tat\_Cyrl&0.593&0.529&0.637&0.564&0.642&0.613\\
&kaz\_Cyrl&0.632&0.495&0.598&0.559&0.603&0.627\\
&urd\_Arab&0.598&0.441&0.603&0.588&0.657&0.676\\
\hline
LLaMA-3.2&dzo\_Tibt&0.250&0.113&0.480&0.627&0.206&0.235\\
&nqo\_Nkoo&0.417&0.201&0.523&0.730&0.377&0.333\\
&sat\_Olck&0.505&0.176&0.593&0.754&0.422&0.368\\
&taq\_Tfng&0.475&0.181&0.235&0.225&0.338&0.260\\
&tir\_Ethi&0.387&0.103&0.373&0.417&0.343&0.368\\
&min\_Arab&0.451&0.186&0.537&0.726&0.407&0.255\\
&nus\_Latn&0.382&0.186&0.319&0.422&0.436&0.348\\
&kac\_Latn&0.539&0.196&0.206&0.392&0.559&0.319\\
&luo\_Latn&0.500&0.250&0.373&0.520&0.485&0.436\\
&ayr\_Latn&0.363&0.255&0.446&0.426&0.412&0.343\\
&fuv\_Latn&0.426&0.221&0.319&0.431&0.475&0.387\\
&wol\_Latn&0.456&0.275&0.343&0.412&0.539&0.422\\
&ckb\_Arab&0.529&0.333&0.657&0.515&0.608&0.485\\
&khk\_Cyrl&0.490&0.255&0.505&0.598&0.603&0.534\\
&tgk\_Cyrl&0.471&0.270&0.578&0.672&0.618&0.510\\
&azb\_Arab&0.515&0.279&0.495&0.461&0.627&0.510\\
&eus\_Latn&0.515&0.324&0.623&0.554&0.588&0.627\\
&tat\_Cyrl&0.593&0.309&0.618&0.657&0.711&0.632\\
&kaz\_Cyrl&0.500&0.348&0.691&0.657&0.735&0.676\\
&urd\_Arab&0.588&0.284&0.637&0.431&0.662&0.593\\
\hline
Gemma-2 &nqo\_Nkoo&0.417&0.137&0.696&0.671&0.255&0.402\\
&dzo\_Tibt&0.250&0.127&0.578&0.569&0.240&0.230\\
&taq\_Tfng&0.475&0.167&0.240&0.230&0.353&0.431\\
&nus\_Latn&0.382&0.206&0.446&0.338&0.338&0.382\\
&min\_Arab&0.451&0.216&0.672&0.614&0.368&0.407\\
&kac\_Latn&0.539&0.328&0.436&0.314&0.436&0.520\\
&ayr\_Latn&0.363&0.270&0.402&0.417&0.363&0.402\\
&sat\_Olck&0.505&0.240&0.686&0.622&0.480&0.549\\
&tir\_Ethi&0.387&0.314&0.466&0.314&0.446&0.500\\
&fuv\_Latn&0.426&0.333&0.485&0.358&0.500&0.520\\
&luo\_Latn&0.500&0.328&0.569&0.377&0.480&0.515\\
&wol\_Latn&0.456&0.412&0.505&0.363&0.529&0.603\\
&ckb\_Arab&0.529&0.333&0.618&0.554&0.559&0.564\\
&tgk\_Cyrl&0.471&0.407&0.696&0.593&0.608&0.593\\
&khk\_Cyrl&0.490&0.368&0.637&0.539&0.554&0.578\\
&azb\_Arab&0.515&0.485&0.603&0.471&0.642&0.667\\
&tat\_Cyrl&0.593&0.466&0.686&0.598&0.735&0.721\\
&eus\_Latn&0.515&0.495&0.711&0.613&0.716&0.711\\
&urd\_Arab&0.588&0.495&0.691&0.480&0.779&0.765\\
&kaz\_Cyrl&0.500&0.515&0.667&0.657&0.740&0.740\\

\hline
\end{tabular}
}
\caption{Zero-shot ICL with language alignments and few-shot ICL with or without alignment over SIB-200 on DeepSeek, LLaMA-3.2 and Gemma-2.}
\label{tab:PEFT zero-shot results}
\end{table*}

\end{document}